\documentclass[10pt,logo,copyright]{nvidia_release_template/enpire_nvidia_report}
\usepackage[authoryear,sort&compress,round]{natbib}
\setcitestyle{numbers}
\usepackage{colortbl}
\usepackage{xspace}
\usepackage{enumitem}
\usepackage{amsmath,amssymb}
\usepackage{booktabs}
\usepackage{longtable}
\usepackage{array}
\usepackage{algpseudocode}
\usepackage{algorithm}
\usepackage{titletoc}
\usepackage[frozencache,cachedir=minted-cache]{minted}
\usepackage{listings}
\usepackage{xltabular}
\usepackage[nameinlink]{cleveref}
\usepackage{caption}

\graphicspath{{./}{corl_2026_template_submission/}{corl_2026_template_submission/figures/}{corl_2026_template_submission/figures/example/}}

\usepackage{tcolorbox}
\usepackage{minted}
\input{nvidia_release_template/JetBrainsMono-Font-TTF/t1JetBrainsMono.fd}
\pdfmapline{+JetBrainsMono_Rg < nvidia_release_template/JetBrainsMono-Font-TTF/JetBrainsMono-Regular.ttf <T1-WGL4.enc}
\pdfmapline{+JetBrainsMono_Bd < nvidia_release_template/JetBrainsMono-Font-TTF/JetBrainsMono-Bold.ttf <T1-WGL4.enc}
\pdfmapline{+JetBrainsMono_It < nvidia_release_template/JetBrainsMono-Font-TTF/JetBrainsMono-Italic.ttf <T1-WGL4.enc}
\pdfmapline{+JetBrainsMono_BdIt < nvidia_release_template/JetBrainsMono-Font-TTF/JetBrainsMono-BoldItalic.ttf <T1-WGL4.enc}

\tcbuselibrary{minted,skins,breakable}
\usepackage{mathrsfs}
\usepackage{amsmath}
\usepackage{physics}
\usepackage{mathtools}
\usepackage{dsfont} 
\usepackage{float}

\usepackage{subcaption}
\usepackage{graphicx}
\usepackage{geometry} 

\usepackage{booktabs} 
\usepackage{multirow} 
\usepackage{array}    
\usepackage{tcolorbox}

\usepackage{makecell} 

\definecolor{codeBg}{HTML}{F7F8FA}
\definecolor{codeFrame}{HTML}{D9DEE7}
\definecolor{codeTitleBg}{HTML}{2F3747}
\definecolor{codeTitleText}{HTML}{FFFFFF}
\definecolor{codeLine}{HTML}{7A8494}
\definecolor{codeKeyword}{HTML}{0000FF}
\definecolor{codeString}{HTML}{A31515}
\definecolor{codeComment}{HTML}{008000}
\definecolor{codeNumber}{HTML}{098658}
\definecolor{codeType}{HTML}{267F99}
\definecolor{codeFunc}{HTML}{795E26}
\makeatletter
\newcommand{\appendixcodefontsize}{%
  \footnotesize
  \fontsize{0.8\dimexpr\f@size pt\relax}{0.96\dimexpr\f@size pt\relax}\selectfont
}
\makeatother

\makeatletter
\newcommand{\ApplyCodePalette}{%
  \@namedef{PYG@tok@c}{\def\PYG@tc####1{\textcolor{codeComment}{####1}}}%
  \@namedef{PYG@tok@ch}{\def\PYG@tc####1{\textcolor{codeComment}{####1}}}%
  \@namedef{PYG@tok@cm}{\def\PYG@tc####1{\textcolor{codeComment}{####1}}}%
  \@namedef{PYG@tok@cp}{\def\PYG@tc####1{\textcolor{codeComment}{####1}}}%
  \@namedef{PYG@tok@cpf}{\def\PYG@tc####1{\textcolor{codeComment}{####1}}}%
  \@namedef{PYG@tok@c1}{\def\PYG@tc####1{\textcolor{codeComment}{####1}}}%
  \@namedef{PYG@tok@cs}{\def\PYG@tc####1{\textcolor{codeComment}{####1}}}%
  \@namedef{PYG@tok@k}{\let\PYG@bf\textbf\def\PYG@tc####1{\textcolor{codeKeyword}{####1}}}%
  \@namedef{PYG@tok@kc}{\let\PYG@bf\textbf\def\PYG@tc####1{\textcolor{codeKeyword}{####1}}}%
  \@namedef{PYG@tok@kd}{\let\PYG@bf\textbf\def\PYG@tc####1{\textcolor{codeKeyword}{####1}}}%
  \@namedef{PYG@tok@kn}{\let\PYG@bf\textbf\def\PYG@tc####1{\textcolor{codeKeyword}{####1}}}%
  \@namedef{PYG@tok@kp}{\let\PYG@bf\textbf\def\PYG@tc####1{\textcolor{codeKeyword}{####1}}}%
  \@namedef{PYG@tok@kr}{\let\PYG@bf\textbf\def\PYG@tc####1{\textcolor{codeKeyword}{####1}}}%
  \@namedef{PYG@tok@kt}{\def\PYG@tc####1{\textcolor{codeType}{####1}}}%
  \@namedef{PYG@tok@nc}{\let\PYG@bf\textbf\def\PYG@tc####1{\textcolor{codeType}{####1}}}%
  \@namedef{PYG@tok@nb}{\def\PYG@tc####1{\textcolor{codeType}{####1}}}%
  \@namedef{PYG@tok@nf}{\def\PYG@tc####1{\textcolor{codeFunc}{####1}}}%
  \@namedef{PYG@tok@fm}{\def\PYG@tc####1{\textcolor{codeFunc}{####1}}}%
  \@namedef{PYG@tok@nd}{\def\PYG@tc####1{\textcolor{codeFunc}{####1}}}%
  \@namedef{PYG@tok@s}{\def\PYG@tc####1{\textcolor{codeString}{####1}}}%
  \@namedef{PYG@tok@sd}{\def\PYG@tc####1{\textcolor{codeString}{####1}}}%
  \@namedef{PYG@tok@s1}{\def\PYG@tc####1{\textcolor{codeString}{####1}}}%
  \@namedef{PYG@tok@s2}{\def\PYG@tc####1{\textcolor{codeString}{####1}}}%
  \@namedef{PYG@tok@sb}{\def\PYG@tc####1{\textcolor{codeString}{####1}}}%
  \@namedef{PYG@tok@se}{\def\PYG@tc####1{\textcolor{codeString}{####1}}}%
  \@namedef{PYG@tok@si}{\def\PYG@tc####1{\textcolor{codeString}{####1}}}%
  \@namedef{PYG@tok@sr}{\def\PYG@tc####1{\textcolor{codeString}{####1}}}%
  \@namedef{PYG@tok@ss}{\def\PYG@tc####1{\textcolor{codeString}{####1}}}%
  \@namedef{PYG@tok@dl}{\def\PYG@tc####1{\textcolor{codeString}{####1}}}%
  \@namedef{PYG@tok@m}{\def\PYG@tc####1{\textcolor{codeNumber}{####1}}}%
  \@namedef{PYG@tok@mi}{\def\PYG@tc####1{\textcolor{codeNumber}{####1}}}%
  \@namedef{PYG@tok@mf}{\def\PYG@tc####1{\textcolor{codeNumber}{####1}}}%
  \@namedef{PYG@tok@mh}{\def\PYG@tc####1{\textcolor{codeNumber}{####1}}}%
  \@namedef{PYG@tok@mo}{\def\PYG@tc####1{\textcolor{codeNumber}{####1}}}%
  \@namedef{PYG@tok@mb}{\def\PYG@tc####1{\textcolor{codeNumber}{####1}}}%
  \@namedef{PYG@tok@il}{\def\PYG@tc####1{\textcolor{codeNumber}{####1}}}%
}
\makeatother

\DeclareCaptionType{codesnippet}[Code Snippet][List of Code Snippets]
\crefname{codesnippet}{code snippet}{code snippets}
\Crefname{codesnippet}{Code Snippet}{Code Snippets}

\newtcblisting{appendixcode}[2][]{%
  enhanced,
  breakable,
  width=\linewidth,
  listing only,
  listing engine=minted,
  minted language=#2,
  minted options={
    fontsize=\appendixcodefontsize,
    linenos,
    numbersep=6pt,
    xleftmargin=1.8em,
    breaklines,
    autogobble,
    tabsize=4,
    formatcom=\ApplyCodePalette
  },
  colback=codeBg,
  colframe=codeFrame,
  boxrule=0.45pt,
  arc=0mm,
  outer arc=0mm,
  left=0mm,
  right=1mm,
  top=1mm,
  bottom=1mm,
  boxsep=1mm,
  before skip=4pt,
  #1
}

\newcommand{\websiteurl}{enpire-research.github.io}
\newcommand{\website}{\href{https://\websiteurl}{\websiteurl}}

\DeclareMathAlphabet{\mathcal}{OMS}{cmsy}{m}{n}
\SetMathAlphabet{\mathcal}{bold}{OMS}{cmsy}{b}{n}

\newcommand{\argmax}[1]{\mathrm{argmax}}
\newcommand{\argmin}[1]{\mathrm{argmin}}

\usepackage{amsthm}
\theoremstyle{plain}

\theoremstyle{definition}

\theoremstyle{remark}

\usepackage{cancel}
\usepackage{algorithm}
\usepackage{algpseudocode}

\renewcommand{\websiteurl}{research.nvidia.com/labs/gear/enpire} 

\definecolor{nvidiaGreen}{RGB}{118,185,0}
\definecolor{cmuRed}{HTML}{C41230}
\newcommand{\method}{\textbf{\texttt{{ENPIRE}}}\xspace}
\newcommand{\methodcolor}{\textbf{\texttt{{\textcolor{cmuRed}{EN}\textcolor{nvidiaGreen}{PIRE}}}}\xspace}

\crefname{equation}{Eq.}{Eqs.}
\crefname{figure}{Fig.}{Figs.}
\crefname{section}{Sec.}{Secs.}
\crefname{appendix}{App.}{Apps.}
\crefname{table}{Tab.}{Tabs.}
\crefname{algorithm}{Alg.}{Algs.}

\title{\method: Agentic Robot Policy Self-Improvement in the Real World}
\author{
\parbox{\textwidth}{
\raggedright
{\small\bfseries
Wenli Xiao$^{1,2*}$,
Jia Xie$^{2*}$,
Tonghe Zhang$^{2*}$,
Haotian Lin$^{2*}$,
Letian ``Max'' Fu$^{3}$,
Haoru Xue$^{3}$,
Jalen Lu$^{2}$}\\
{\small\bfseries
Yi Yang$^{2}$,
Cunxi Dai$^{2}$,
Zi Wang$^{1}$,
Jimmy Wu$^{1}$,
Guanzhi Wang$^{1}$,
S. Shankar Sastry$^{3}$,
Ken Goldberg$^{3}$}\\
{\small\bfseries
Linxi ``Jim'' Fan$^{1,\dagger}$,
Yuke Zhu$^{1,\dagger}$,
Guanya Shi$^{2,\dagger}$}\\
{\footnotesize
$^{1}$NVIDIA \quad
$^{2}$CMU \quad
$^{3}$UC Berkeley \quad
$^{*}$Equal contribution \quad
$^{\dagger}$Equal advising}
}
}
\newcommand{\nvidiareleaseomitappendixfleet}{}

\begin{document}

\maketitle

\begin{center}
    \includegraphics[width=1.0\textwidth]{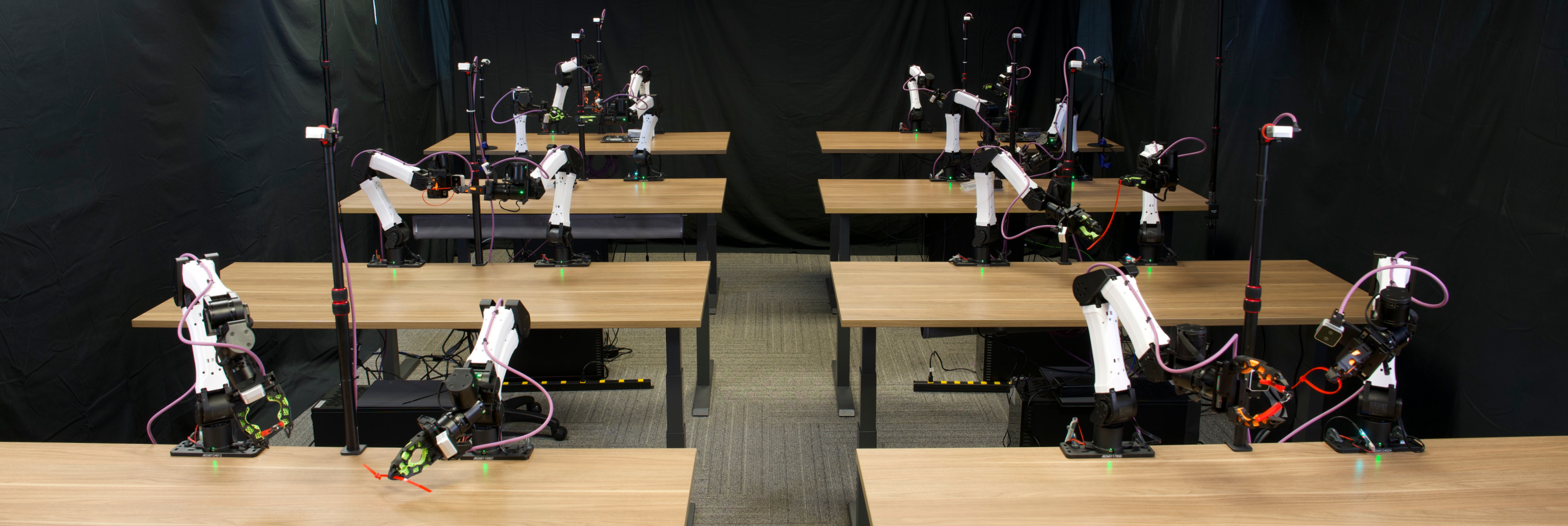}
\end{center}
\vspace{0.3em}

\begingroup
\small
\refstepcounter{figure}\label{fig:fleet}

\vspace{-0.5em}
\noindent\makebox[\linewidth][c]{%
\begin{minipage}{0.9\linewidth}
Figure~\thefigure: \textbf{Robot fleet for physical autoresearch}. The fleet contains eight bimanual YAM robot stations. Each station owns its robot hardware, compute, and coding agent. Website: \website, Codebase: \href{https://github.com/NVlabs/ENPIRE}{github.com/NVlabs/ENPIRE}.
\end{minipage}%
}
\par

\endgroup

\begin{abstract}
Achieving dexterous robotic manipulation in the real world relies heavily on human supervision and algorithmic engineering, which is a central bottleneck in the pursuit of general physical intelligence. Although emerging coding agents can generate code to automate algorithm search, their successes remain largely confined to digital environments. We conjecture that the missing abstraction to automate robotics research is a \emph{repeatable feedback loop} for real-world policy improvement: reset the scene, execute a policy, verify the outcome, and refine the next iteration.
To bridge this gap, we introduce \methodcolor, a harness framework for coding agents that instantiates this physical feedback routine with four core modules: an \underline{En}vironment module (\textbf{\texttt{{\textcolor{cmuRed}{EN}}}}) for automatic reset and verification, a \underline{P}olicy \underline{I}mprovement module (\textbf{\texttt{{\textcolor{nvidiaGreen}{PI}}}}) that launches policy refinement, a \underline{R}ollout module (\textbf{\texttt{{\textcolor{nvidiaGreen}{R}}}}) to evaluate policies with single or multiple physical robots operating in parallel, and an \underline{E}volution module (\textbf{\texttt{{\textcolor{nvidiaGreen}{E}}}}) in which coding agents analyze logs, consult literature, improve training infrastructure and algorithm code to address failure modes. This closed-loop system transforms real-world robot learning into a controllable optimization procedure that agents can manage, thus minimizing human effort while allowing fair ablations across training recipes and agent variants. Powered by \method, frontier coding agents can autonomously develop a policy to achieve a 99\% success rate on challenging, dexterous manipulation tasks in the real world, such as PushT, organizing pins into a pin box, and using a cutter to cut a zip tie. Coding agents can improve policies with various \textbf{\texttt{{\textcolor{nvidiaGreen}{PI}}}} regimes, such as heuristic learning, tool calling, behavior cloning, offline or online reinforcement learning. Moreover, \method can be significantly accelerated on a robot fleet, and we propose two metrics, namely, \underline{M}ean \underline{R}obot \underline{U}tilization (MRU) and \underline{M}ean \underline{T}oken \underline{U}tilization (MTU) to measure the efficiency of multi-agent physical autoresearch. We also include simulation results in \texttt{RoboCasa}. Our findings suggest a practical and scalable path towards autonomously advancing robotics in the real world.
\end{abstract}
\abscontent


\begin{figure}
    \centering
    \includegraphics[width=1\linewidth,clip]{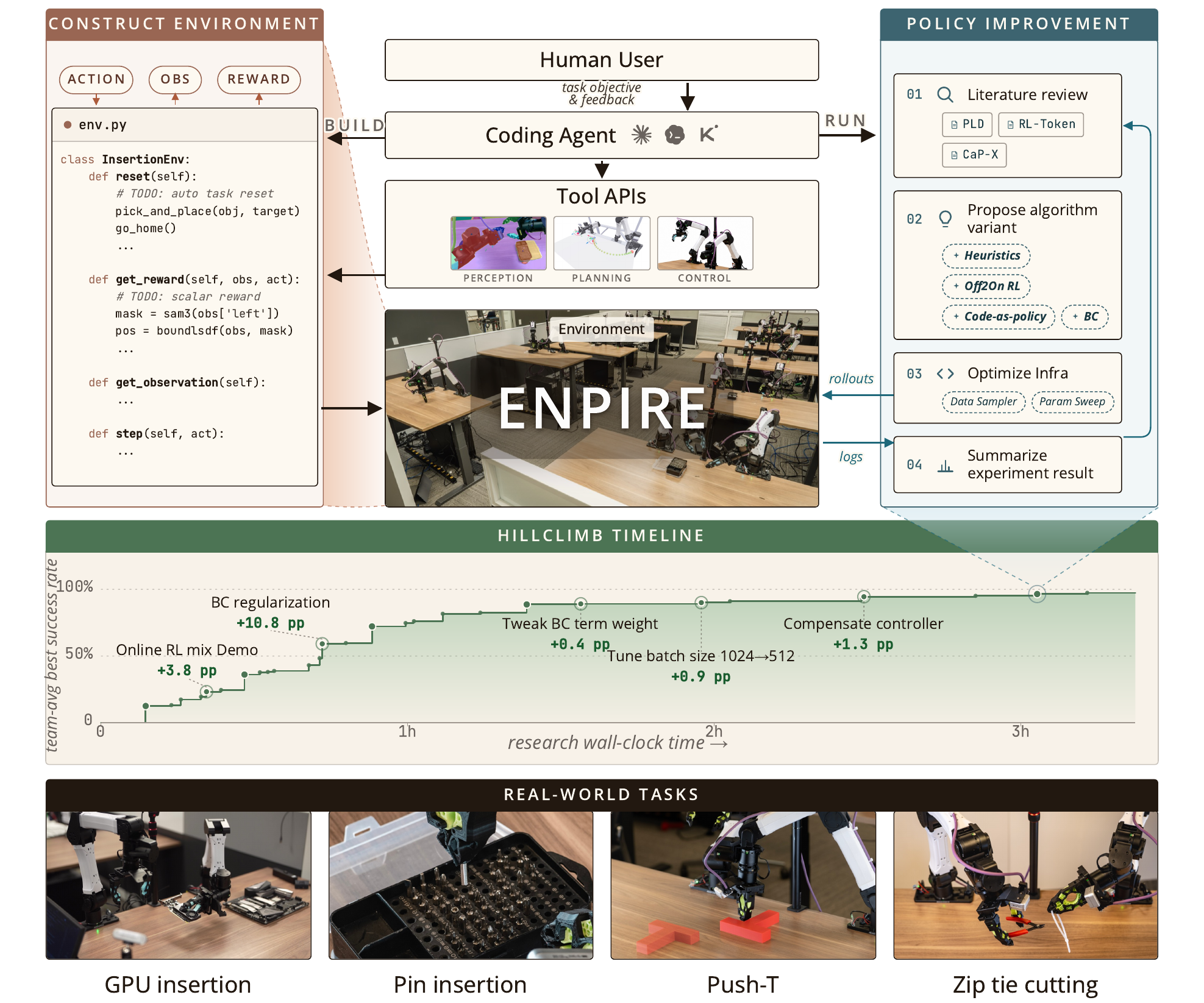}
    \caption{
    Overview of physical autoresearch framework \method. To solve a dexterous task in the real world, \method first uses tool calls to construct an environment with automatic reset and verification mechanisms according to human feedback. Next, the coding agent calls environment APIs to conduct autoresearch and hill-climb the policy success rate from real-world reward signals. More results are available at \website.
    }
    \label{fig:overview}
\end{figure}

\section{Introduction}\label{sec:intro}
Automatically learning dexterous manipulation skills has been a major obstacle in the journey towards general physical intelligence. Frontier policy training methods, although highly effective, still rely on humans behind the scenes to participate in the data collection, evaluation with reset, and algorithm adjustment life cycle~\citep{pi05, recap, rl100}. As we scale up real-world policy learning,  humans labor of babysitting policy improvement inevitably become one of the limiting factors of how quickly robots acquire dexterity. 

Recent advances in autonomous research~\citep{karpathy2025autoresearch} (autoresearch), powered by coding agents, show a promising path toward automating algorithmic improvements. However, unique challenges arise when we extend this paradigm to automate real-world policy learning. First, coding agents lack a set of \emph{real-world environment interfaces} for closed-loop hypothesis testing in the physical world, which requires automated policy deployment, evaluation, and scene resetting. Second, when \emph{scaling up} autoresearch throughput across robot fleet, it remains an open problem to select and verify hypotheses while maintaining efficiency of resource utilization. 

To address these challenges, we introduce \method, an agent harness framework that leverages a suite of autonomous environment interfaces to enable scalable physical autoresearch. \method decomposes the acquisition of dexterous manipulation skills into two autoresearch procedures. In the first phase (\textbf{\texttt{{EN}}} in \method), coding agents construct an autonomous environment interface from human feedback. Through this initial research loop, agents implement and optimize procedural tool calls that establish safety boundaries, automated reset, and verification procedures for a specific task. They are optimized offline with a one-time setup cost; once finalized, they serve as immutable APIs that are reused throughout the subsequent stage. The second phase (\textbf{\texttt{{PIRE}}} in \method) transitions to a fully autonomous autoresearch procedure, where coding agents refine policies based on real-world feedback. Guided entirely by the environment's automated verification signals, the agents independently explore and optimize diverse methodologies to maximize real-world task success rates without human intervention. 

To scale this pipeline across multiple robots in parallel, \method introduces a mechanism to evolutionarily select hypotheses, which accelerates policy improvement. We dispatch a decentralized team of agents to test training recipes asynchronously, sharing and abandoning ideas based on the average success rates. The accumulated knowledge can be transferred to similar, novel tasks. 

\noindent We highlight the contributions of this work as follows:
\begin{itemize}[leftmargin=*, topsep=2pt, itemsep=2pt, parsep=0pt, partopsep=0pt]
  \item We formalize physical autoresearch for dexterous manipulation as a distinct problem for the coding agent and robotics communities. We propose to tackle this with a two-step approach: first, conducting a human-guided autoresearch to construct automatic environment feedback and verify it offline, then executing a fully autonomous autoresearch from real-world feedback in an online fashion to improve the policy. 
  \item We implement \method, an agentic harness through which a coding agent develops reusable robotic tools, constructs rewards and reset mechanisms with procedural tool calls, and effectively optimizes learning algorithms on real robots without human intervention. 
  \item We demonstrate that \method can hill-climb the success rate to a high level in multiple dexterous manipulation tasks. In pin insertion, policy convergence to 100\% is achieved faster than a frontier human-in-the-loop method~\citep{pld}. We also observe initial scaling behavior for parallelized autoresearch on a robot fleet and propose key metrics, Mean Token Utilization (MTU) and Mean Robot Utilization (MRU), to benchmark the efficiency of this procedure. 
\end{itemize}

\ifdefined\nvidiareleaseomitsharedteaser
\else
\begin{figure}
    \centering
    \includegraphics[width=0.8\linewidth]{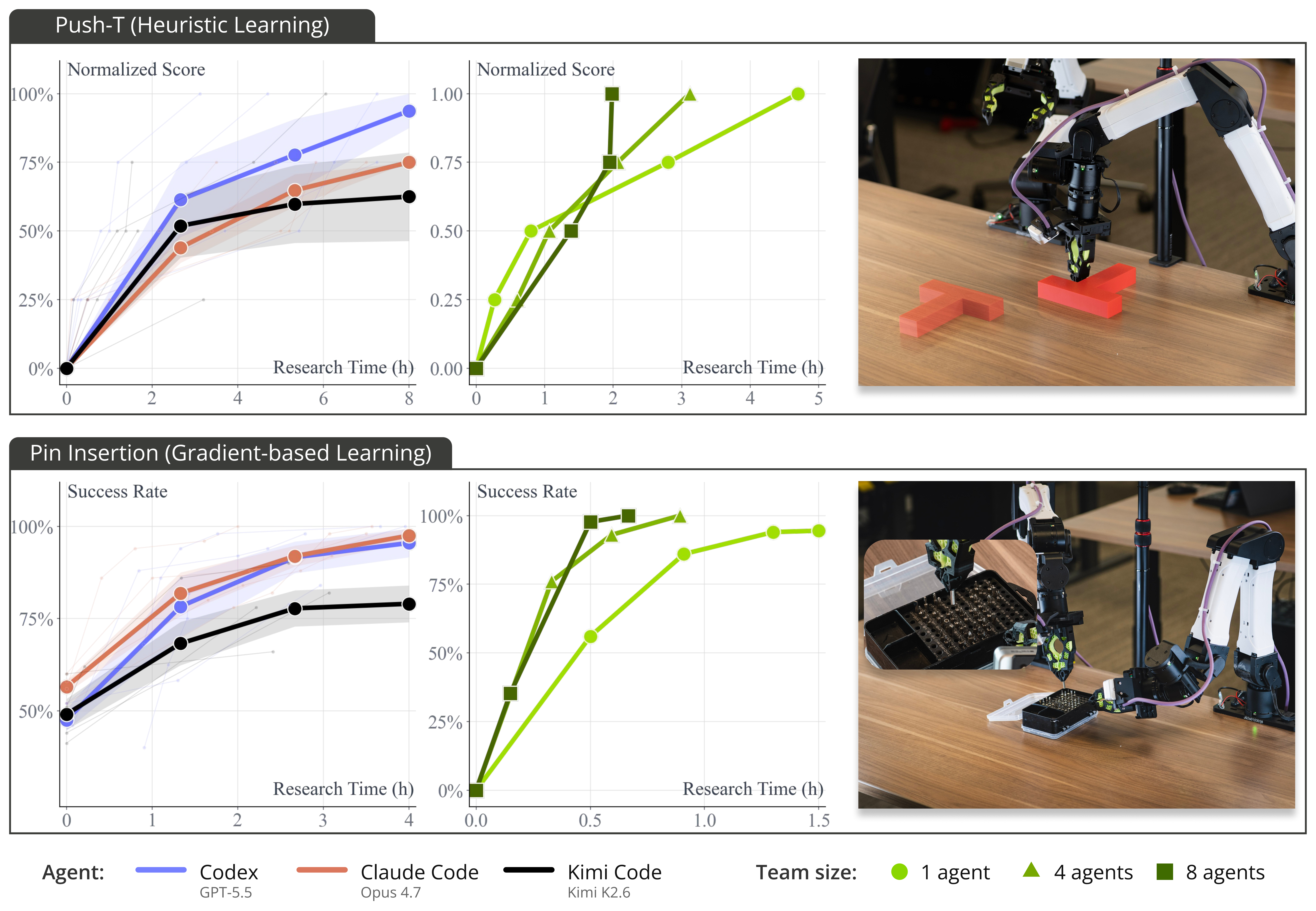}
    \caption{\textbf{Benchmarking coding agents for physical autoresearch}. \method enables \texttt{state-of-the-art} coding agents to achieve autonomous policy improvement on \textbf{Push-T} and \textbf{Pin Insertion} tasks. \method also scales with resources, as increasing the number of robot agent workers reduces the wall-clock time required to reach the same task performance.} 
    \label{fig:real_pusht_pin_insertion}
\end{figure}
\fi


\section{Method}\label{sec: method}
In this section, we present a harness design that allows coding agents to perform automated research in the physical world. This procedure is decomposed into two stages: \emph{environment construction from human feedback} (\textbf{\texttt{{EN}}} in \method) and \emph{automatic policy improvement from real-world feedback} (\textbf{\texttt{{PIRE}}} in \method). 


\subsection{Stage One: Environment Construction from Human Feedback}\label{sec:method-env}
For coding agents to conduct autonomous physical autoresearch, we need to first construct an agent-friendly abstraction of physical interaction and feedback as a structured environment. This includes task-specific safety constraints to support long-term operation, a real-time automated verification process for feedback and credit assignment, and a robust automatic reset mechanism to ensure fast iteration. To improve reliability, agents construct the environment APIs with procedural tool calls and refine their implementation according to human assessment. Human effort is a one-time cost that will be amortized across all implementations on all robots during subsequent automatic policy improvement in \cref{sec:method-tree}. 

\paragraph{Hard safety constraints} We restrict the configuration space and kinematic behaviors of the robot to a safe operational limit. The safety regions are sufficient for task completion and serve as hard constraints: violating these limits triggers an immediate task failure and an automated reset. This serves as a safeguard for real-world interaction, also as a source of episode termination or truncation.

\begin{figure}[H]
    \centering
    \includegraphics[width=0.95\linewidth]{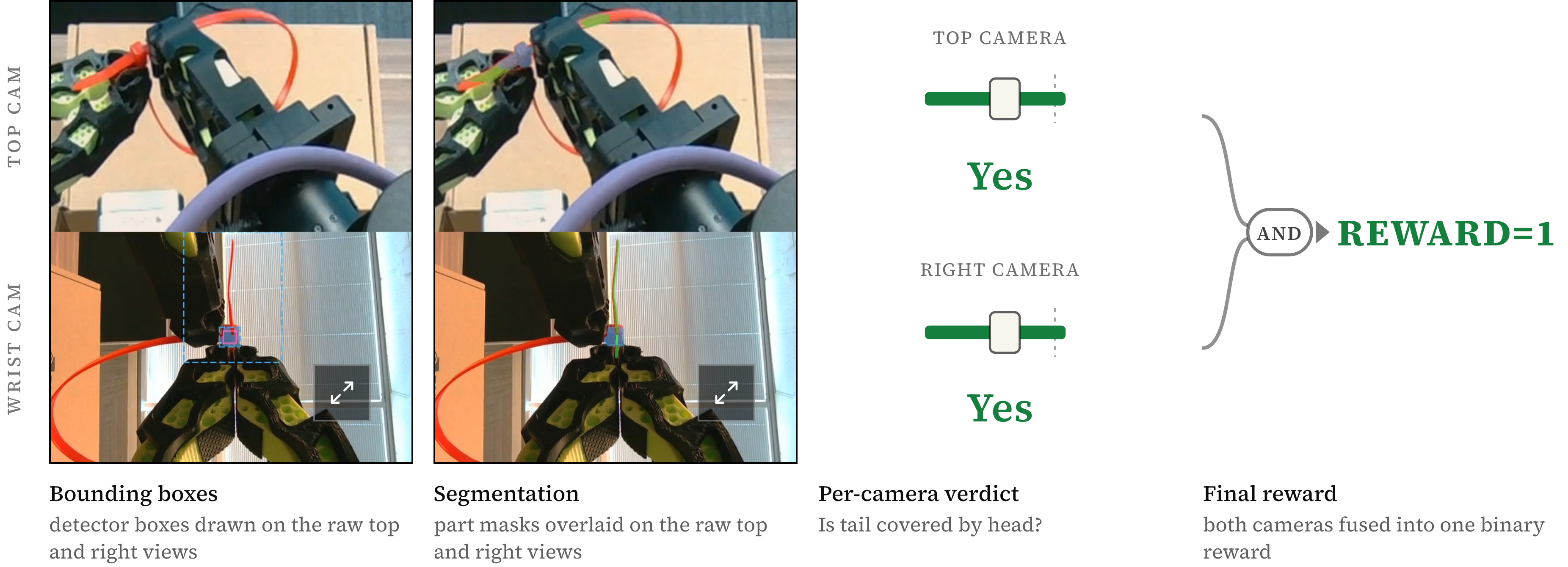}
    \caption{Reward for zip-tie insertion. Cropping and image segmentation test whether the zip-tie strap passes through the zip-tie head. Two camera views are considered to prevent false positives.}
    \label{fig:zip_tie_reward}
\end{figure}

\paragraph{Automated verification} 
During real-world experiments, robot learning needs real-time verification from sensor inputs to quantify experimental progress. To minimize human engineering efforts, coding agents are tasked with synthesizing a binary reward function from procedural tool calls to distinguish task outcomes. Given only a few minutes of success and failure demonstrations, the agents use the videos and proprioception recordings to maximize the prediction accuracy score while minimizing processing latency. For example, autoresearch discovers a robust reward function for pin insertion that is based on visual alignment, end-effector height, and force estimates. 
Coding agents also show their ability to design perception-dependent rewards in zip-tie insertion in \cref{fig:zip_tie_reward} and optimize its inference latency to under 150ms, which approaches the reactiveness of the human visual system~\citep{Thorpeetal1996}. More details are provided in the appendix. 

\paragraph{Automated reset}
Once a task is completed or fails, the agent executes a series of tool calls to restore the environment to its initial state. For contact-rich tasks, we employ procedural tool calls inspired by CaP-X~\citep{capx2026}, using modular manipulation skills to reset the environment directly to the start of the most challenging phase. By positioning the robot at the precise onset of critical actions, such as inserting a pin, seating a GPU, or grasping scissors, we strategically focus the learning system on these precision bottlenecks.

The safety constraints, automated verification, and automated reset constitute our Environment module (\textbf{\texttt{{EN}}} in \method). Policies also receive visual-proprioception sensor inputs and submit action commands to the robot controller, which, combined with our reward, constitutes the Rollout module (\textbf{\texttt{{R}}} in \method). Once constructed, coding agents have access to these modules through immutable Gym APIs~\citep{gym}, which provide feedback signals and debugging information for automated policy improvement. 

\subsection{Stage Two: Automated Policy Improvement from Real-World Feedback}\label{sec:method-tree}
The second stage leverages automated research to train a policy for a specific task. Upon initialization, the coding agent receives the task description with the ultimate objective of maximizing the task's success rate through autonomous experimentation. To achieve this, the agent is granted write permissions to a streamlined training codebase that supports basic end-to-end policy training and code-based policy synthesis. 
\vspace{-2mm}
\paragraph{A motivating example}
We illustrate this process using a pin insertion task, where the policy must insert a pin into a hole with a tight 4mm clearance (\cref{fig:overview}). During this automated research phase, the agent operates through a Policy Improvement module (\textbf{\texttt{{PI}}} in \method), where it reviews the literature to generate insights, formulates hypotheses, and directly modifies the training code---such as behavior cloning or RL algorithms---to optimize performance based on real-world automatic verification results. To gather rich evidence, the agent invokes environment APIs to log robot trajectories, video recordings, and reward signals during rollouts, and inspects these statistics to guide continued improvement. 

\vspace{-2mm}
\paragraph{Accelerating autoresearch via multi-agent scaling}\label{sec:method-fleet}
While automatic reset and verification mechanisms enhance the scalability of a single deployment loop, physical autoresearch can be further accelerated by launching a parallel, multi-agent decentralized collaboration protocol. This protocol deploys $N$ agents across $N$ physical robots to test $N$ hypotheses asynchronously. Each agent branches off from the same baseline policy training codebase, collaborating autonomously via Git to scale this Evolution module (\textbf{\texttt{{E}}} in \textbf{\texttt{{PIRE}}}). Without human intervention, agents spontaneously cherry-pick, copy, or merge successful training recipes from their peers to optimize their code search. Empirically, we observe that scaling the number of parallel agent-robot pairs drastically reduces the wall-clock time required to discover a high-success-rate policy recipe, as illustrated in Fig.~\ref{fig:real_pusht_pin_insertion}. 

To quantify how efficiently a coding agent converts its allocated physical resources into research progress, we propose two utilization metrics that complement task-level outcomes. Mean Robot Utilization (MRU) is the fraction of research wall-clock time during which the robot is actively executing an experiment. GPU utilization is the fraction of research wall-clock time during which the GPU is actively in use. 
A perfectly resource-saturated agent would push both metrics toward $1$; in practice, neither metric reaches this value for any frontier coding agent we evaluate. 
To measure how efficiently an agent team converts tokens into successful robot policies, we define Mean Token Utilization (MTU) as the fleet's average token consumption throughout autoresearch. We then calculate the token-to-success ratio to reflect the token efficiency of task learning. An empirical result of how robot and token utilization changes with agent team size is illustrated in Fig.~\ref{fig:utilization}.

\section{Experiment}
We show that \method is a capable and scalable agentic autoresearch framework. 
\begin{itemize}[leftmargin=*, topsep=2pt, itemsep=2pt, parsep=0pt, partopsep=0pt]
    \item \emph{First, \method supports the autonomous policy learning after environment construction. }
    \item \emph{Second, the convergence speed of the success rate \method scales with robot and token resources.}
\end{itemize}
We will focus on the following dexterous tasks, which require precise and reactive control from perceptual feedback: \textbf{Push-T}~\citep{chi2025diffusion}, where the robot uses non-prehensile movement to align a T-shape block to a goal region; \textbf{Pin insertion}, where the robot organizes a pin box by plugging in pins into 4mm-diameter holes; \textbf{GPU-insertion}, where the robot seats GPU chips in thin sockets on the motherboard, and \textbf{Ziptie-cutting}, where the robot grabs and closes scissors to cut the tail of a zip tie. A visualization of these tasks is provided in Fig.~\ref{fig:overview}.

We measure success as the chance of completing a task in one rollout, given a fixed number of retries (here, eight). Unlike measuring success by i.i.d. best-of-N sampling, where each attempt is independent, our retries happen after witnessing the previous failure, implying that our metric captures both precision and in-context recovery in the face of environment uncertainty. We stress that recovery, not just one-shot precision, matters for highly dexterous tasks and reflects the robustness needed for reliable deployment in the real world. 

We use \method to conduct autoresearch in various policy learning paradigms. Coding agents have the freedom to choose various methods and their combinations to solve a task, including end-to-end neural network training such as behavior cloning (BC)~\citep{bc} or real-world reinforcement learning (RL)~\citep{realrl-sac}, as well as gradient-free learning methods such as heuristic learning~\citep{weng2026heuristic} and code-based policy synthesis~\citep{cap}. Our real robot platform is the bimanual 6-DoF YAM robot. We benchmark three coding agents in our physical autoresearch experiments, namely, Codex with GPT-5.5 xhigh~\citep{openai2026codex}, Claude Code with Opus 4.7 High~\citep{anthropic2026claudeopus47}, and Kimi Code with Kimi K2.6 thinking~\citep{moonshot2026kimicode}.

\subsection{Autoresearch for Heuristic Learning}
\begin{figure}
    \centering
    \includegraphics[width=0.9\linewidth]{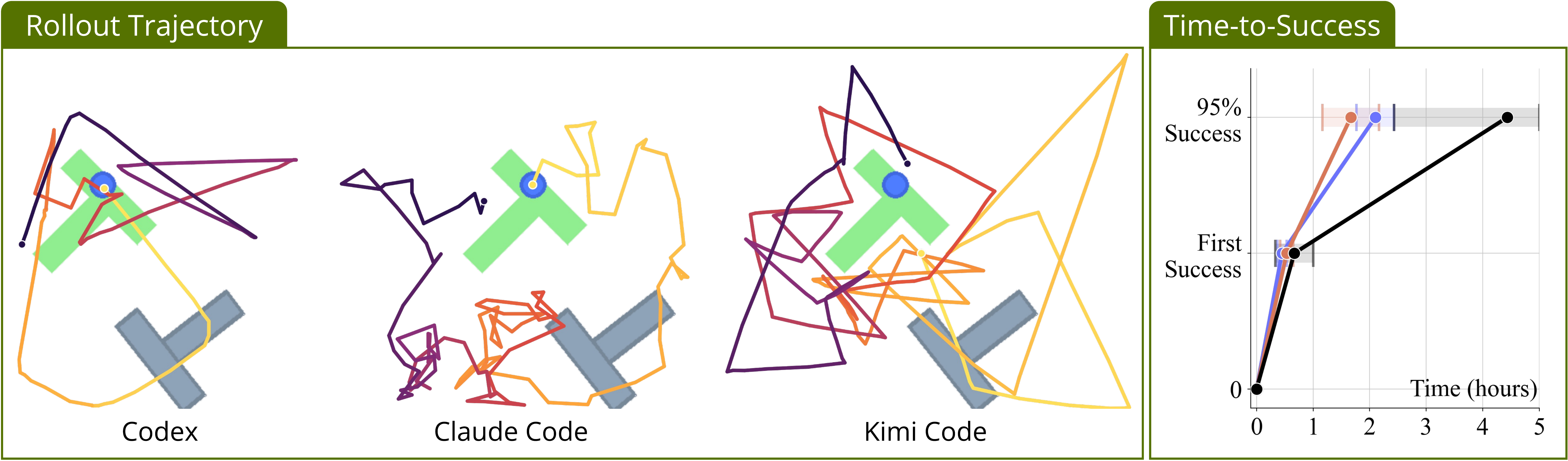}
    \caption{\textbf{Autonomous heuristic learning in simulation}. On the Gym-PushT~\citep{gympusht} benchmark, Claude Code (orange) and Codex (blue) achieve 95\% success rate within approximately 2 hours, while Kimi Code (black) takes twice the time. 
    } 
    \label{fig:sim_real_pusht}
\end{figure}
We tested \method's ability to drive automatic policy improvement in its simplest form:  learning heuristics~\citep{weng2026heuristic} by synthesizing perception and control tool calls~\citep{cap}. To this end, we built a real-world \textbf{Push-T} environment with its simulation counterpart for comparison. 

\paragraph{The unique challenge of physical autoresearch} As illustrated in Figs. \ref{fig:sim_real_pusht} and \ref{fig:real_pusht_pin_insertion}, all coding agents successfully solve the Push-T task in simulation using heuristic learning. However, the real-world environment proves significantly more challenging, causing two of the three agents to fail. While simulators offer consistent, predictable physics for low-variance hypothesis testing, real-world conditions are non-deterministic and time-varying: factors such as robot dynamics, contact friction, and object movements are inherently less predictable and vary across trials and hardware.

To improve real-world robustness, in subsequent tasks, we encourage agents to explore and even combine diverse learning methods that span heuristics and gradient-based learning, to tackle the corner cases in real-world deployment. We also propose scaling up a physical agent team to test hypotheses in diverse and nondeterministic real-world physics. We will elaborate on these designs in subsequent sections. 

\subsection{Autoresearch for Gradient-Based Policy Improvement}
\label{sec: pin insertion}
Apart from supporting heuristic learning, \method is also capable of training end-to-end policies in a precision-critical task, pin insertion, as shown in Fig.~\ref{fig:real_pusht_pin_insertion}. In this task, coding agents are required to achieve 50 consecutive successes in real-world evaluations. The agents tested multiple methods to improve the policy, including behavior cloning (BC), iterative BC with online rollout data aggregation, and online, offline, and offline-to-online RL with a BC regularization term. The agents also tuned parameters such as batch size, the actor-critic policy update rate, and the BC-term hyperparameters.

\subsection{Scaling Policy Learning on a Robot Fleet}
We further study whether scaling the number of robots and agents reduces the wall-clock time required to achieve the same target task performance. We assign one coding agent to one robot, and set up identical environment interfaces, reward functions, and reset mechanisms on this robot fleet. We consider fleet sizes of 1, 4, and 8 robots in Push-T and pin insertion tasks. 

In Push-T, scaling from one to eight agents reduces the time to reach a 1.0 normalized score from roughly five to two hours. In pin insertion, scaling from one to eight agents reduces the time to reach a near-perfect success rate from more than 1.5 hours to approximately 40 minutes.
This shows that \method is capable of transferring additional robot resources to faster policy improvement through distributed hypothesis selection. 


In a multiagent setting, \method can also leverage code-based policies to automatically apply domain randomization during reset. For example, the variations in spatial configurations during GPU insertion span a significantly broader range than prior work~\citep{pld}, which enforces stronger policy robustness.

\begin{figure}
    \centering
    \includegraphics[width=1\linewidth]{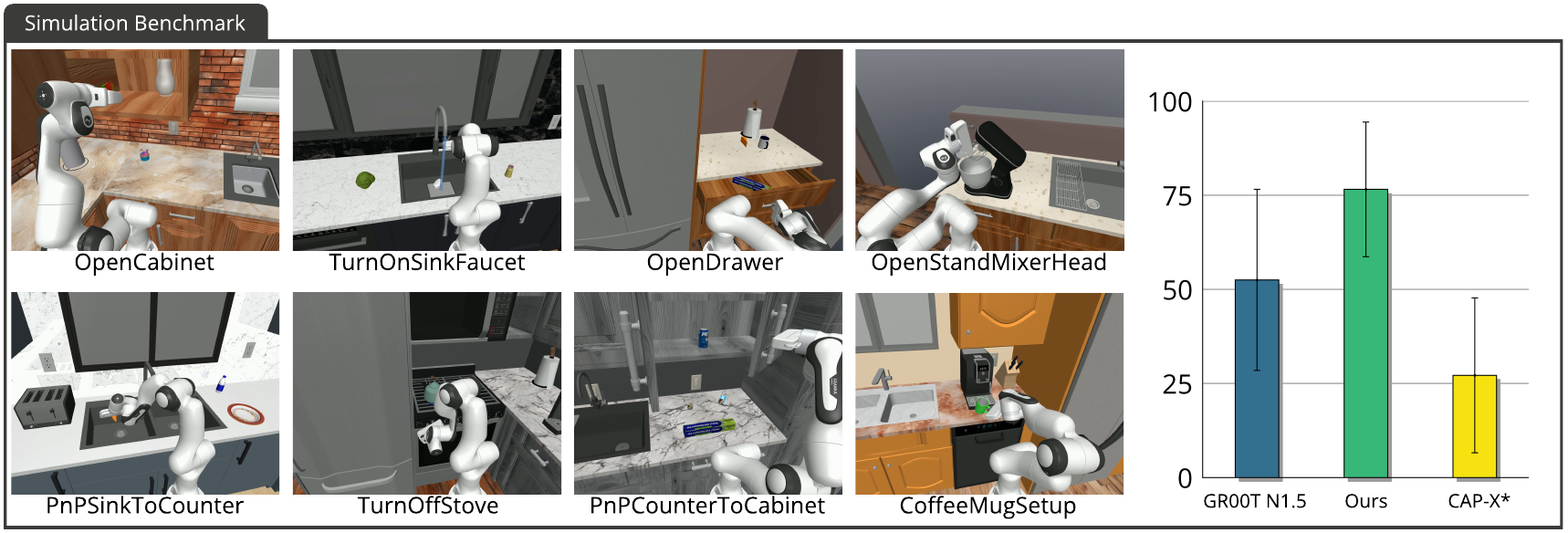}
    \caption{\textbf{Simulation results}. On RoboCasa365~\citep{nasiriany2026robocasa365} benchmark, \method outperforms end-to-end VLA (GR00T~\citep{gr00tN1}) and zero-shot agentic tool use without autoresearch (CaP-X~\citep{capx2026}).}
    \label{fig:placeholder}
\end{figure}

\subsection{Transferring Autoresearch Experience through Agentic Continue Learning} 
We observe that the insights accumulated by multi-agent physical autoresearch are also transferable to similar, novel dexterous tasks. Following the autonomous exploration phase in pin insertion, agents are prompted to document and reflect on the evolution of their training recipes. When we instantiate a new round of autoresearch for the GPU insertion task, appending this knowledge to the new task's instructions allows coding agents to achieve a high success rate. A detailed analysis is provided in the Appendix.

\subsection{\method Discovers Synergy between Code-Based Policies and VLAs} 
Beyond end-to-end or heuristic training, \method can automatically integrate vision-language-action models (VLAs)~\citep{rt1} with procedural tool calls for long-horizon manipulation. In the RoboCasa365 simulator~\citep{nasiriany2026robocasa365}, the agent boosted the success rate of the GR00T VLA~\citep{gr00tN1} by using motion planning and detection tools to hover above an object before grasping. We successfully transferred this strategy to the real world, where the agent learned to hover over scissors, grasp them, and cut a zip tie, as shown in~\Cref{fig:overview}.

\subsection{Quantifying Agent Resource Utilization}
\begin{figure}
    \centering
    \includegraphics[width=1\linewidth]{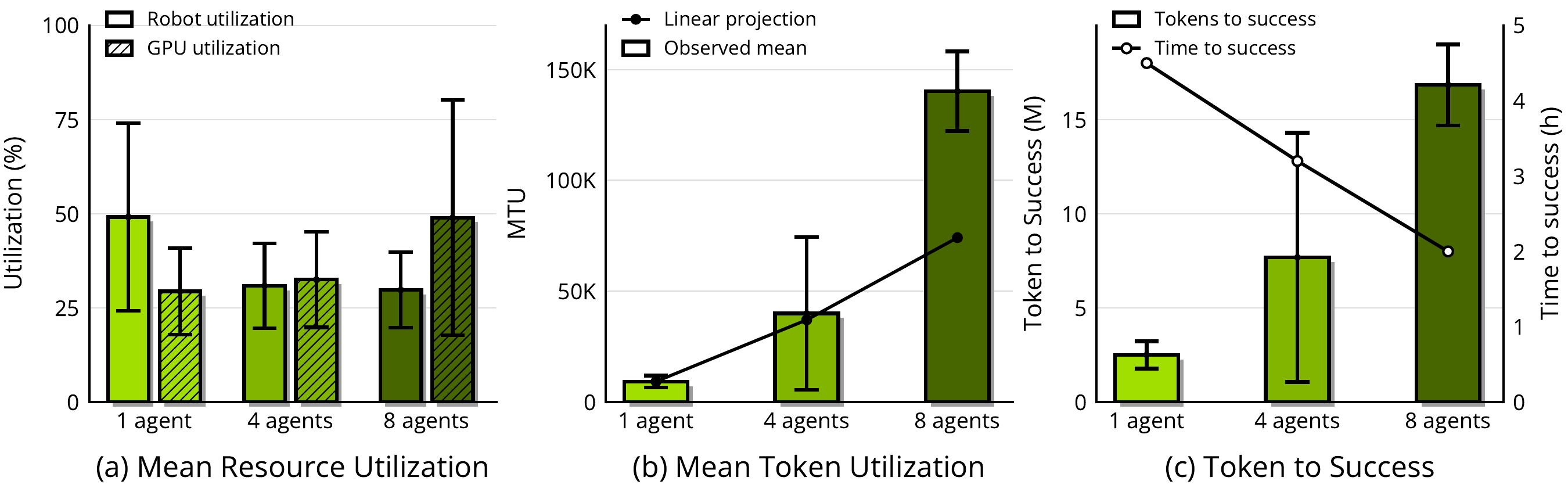}
    \caption{\textbf{Quantifying agent resource utilization}. Scaling from 1 to 8 agents raises GPU utilization and token consumption but lowers per-robot utilization.}
    \label{fig:utilization}
\end{figure}


\Cref{fig:utilization} summarizes the resource-scaling behavior of ENPIRE on the pin insertion task across single-, four-, and eight-agent fleets. We report Mean Robot Utilization (MRU) and GPU utilization using the definitions introduced above. In addition, we measure Mean Token Utilization (MTU), defined as the number of tokens consumed per minute. We further measure Tokens to Success and Time to Success, which quantify the token budget and wall-clock time required to complete the autoresearch objective.

\section{Limitations}
\paragraph{Robot and compute resources are underutilized}

Coding agents do not fully utilize robot resource when they are reading logs, writing code, debugging, or waiting for the language-model backbone. Moreover, as we scale the number of robots, MRU decreases, while GPU active utilization increases (\Cref{fig:utilization}a). Compared to a single-robot setup, the composition of agent activity causes agent teams to spend more of their time summarizing peer branches and less time actually operating the robot. Coding agents may also fail to launch parallel training sessions to exhaust GPU resources.

\paragraph{Token cost grows super-linearly with fleet size.}
As the fleet size increases, token usage grows faster than the ideal linear trend. MTU remains close to the linear projection up to four agents, but rises sharply at eight agents (\Cref{fig:utilization}b). The total token budget required to obtain a successful policy shows the same trend, increasing much more rapidly than the corresponding reduction in wall-clock time (\Cref{fig:utilization}c). Thus, increasing the fleet size trades token efficiency for faster policy improvement: larger fleets reach success sooner, but require a disproportionately higher token budget.

\section{Related Work}\label{sec:related}
\subsection{Coding Agents \& Code as Policies}
\label{sec:related-coding}
This line of work rests on a central abstraction---\emph{executable code as the agent's action}: the model generates a program, the runtime returns structured feedback, and the loop iterates. Code-as-Policies \citep{cap} and ProgPrompt \citep{singh2022progprompt} introduced this formulation for robotics, composing perception and skill APIs into a task plan, and visual reasoning carried the same compositional idea to vision modules \citep{suris2023vipergpt}. Subsequent research extended single-shot generation into multi-turn loops driven by execution feedback---reason-and-act traces \citep{yao2022react}, self-critique and iterative repair \citep{shinn2023reflexion, madaan2023self}, inference-time search \citep{yao2023tree}, and learned tool invocation \citep{schick2023toolformer}; \citet{wang2024codeact} argues code is a stronger action representation than language because each step is runtime-verifiable. A parallel thread formalizes the \emph{agent--computer interface}, a sandboxed shell in which agents read, run, and debug code, now underlying frontier coding products \citep{openai2026codex, anthropic2026claudeopus47, moonshot2026kimicode}.

Within robotics, early code-as-policy systems exposed high-level, human-engineered APIs---closed-form skills, language-conditioned grasping primitives, and affordance scorers \citep{saycan}---leaving the agent to decompose tasks rather than synthesize low-level control. \citet{wang2023voyager} grows a reusable Minecraft skill library via self-verification and an automatic curriculum where trials are near-free, and \citet{capx2026} finds that multi-turn feedback, skill synthesis, and ensembled sampling improve manipulation reliability over low-level primitives. A related line uses LLMs to synthesize auxiliary training signals: reward functions \citep{ma2024eureka, xie2024text2reward, yu2023language}, sim-to-real transfer protocols \citep{ma2024dreureka}, simulation environments \citep{wang2023robogen}, and data-collection pipelines \citep{ahn2024autort}.


\subsection{Agentic Self-Improvement}\label{sec:related-selfimprove}
A self-improvement loop needs each trial to be cheap enough to repeat at scale; systems differ mainly in what a trial returns to the agent. \citet{ellis2021dreamcoder} set the retention pattern that the rest build on: each solved synthesis task adds named subroutines to a growing library, which is viable because execution is free. \citet{wang2023voyager} carried this into an embodied setting, swapping the success signal for LLM self-verification and an automatic curriculum---again viable because Minecraft rollouts cost nothing. \citet{yu2023language}, \citet{ma2024eureka}, and \citet{xie2024text2reward} return a reward rather than a skill: an LLM proposes a dense reward, a policy is trained against it, and the reward is revised from training statistics---a loop Eureka closes via thousands of Isaac Gym rollouts per minute. \citet{ma2024dreureka} reaches toward hardware via synthesized domain-randomization parameters, yet still iterates in simulation and deploys only after revision stops; at the experimental-setting level, \citet{wang2023robogen} generates new tasks and assets in simulation, and \citet{ahn2024autort} coordinates a mobile-robot fleet for offline data collection rather than within-loop iteration. In every case, the loop closes on a cheap substrate while real-robot execution serves only as a sim-to-real or data target, never as the medium of iteration. We retain these skill-accumulation and reward-generation mechanisms but run the loop directly on hardware, where the binding resource is the agent's robot-access budget, not its compute.

\subsection{Autonomous Research Agents and Scientific Discovery}
\label{sec:related-research} The final body of work automates the research loop itself, best read along \emph{the medium in which experiments are launched}. Prior to LLMs, autonomous systems closed the hypothesis--experiment loop on real laboratory hardware \citep{king2004functional, burger2020mobile}. LLM-era systems instead automate the \emph{digital} loop end to end \citep{lu2024aiscientist, yamada2025aiscientistv2, schmidgall2025agentlab, schmidgall2025agentrxiv}, with a recent strand reaching back into the physical sciences through lab automation \citep{boiko2023autonomous, m2024augmenting} or human-executed wet-lab validation \citep{ghareeb2025robin}. Evaluation tracks this digital emphasis: MLE-bench \citep{chan2025mle} scores ML-engineering agents and resource scaling, while analyses of SWE-bench \citep{jimenez2024swebench} caution that gains can reflect contamination over capability. Two gaps follow: no system autonomously runs and optimizes a physical \emph{robotics} loop under an explicit hardware budget---classical robot scientists fixed the apparatus and never wrote their own tools, while LLM research agents never touch real robots---and no benchmark measures utilization of a scarce physical resource rather than capability or cost-per-paper. Our system addresses both gaps by running and optimizing the experimental loop on real robots and introducing resource-utilization metrics for this budget-bound regime.


\section{Acknowledgement}
We are grateful to many colleagues whose help made this project possible. We thank Jason Liu, Tony Tao, Tairan He, Alex Lin, Jim Yang, Paul Zhou, and Abhi Maddukuri for insightful discussions and feedback; Yide Shentu, Bike Zhang, Angchen Xie, Dvij Kalaria, and Yuqi Xie for their support with the experiments; Lion Park, Matin Furutan, Jeremy Chimienti, Dennis Da, and Tri Cao for fleet operation; and Tri Cao for the demo shots. We also thank the NVIDIA GEAR Team and the CMU LeCAR Lab for their continuous support.

\clearpage
\bibliographystyle{plainnat}
\bibliography{corl_2026_template_submission/example}

\newpage
\appendix

\section*{Appendices}
\startcontents[appendix]

All videos are available at \website

\section*{Appendix Outline}
\printcontents[appendix]{}{1}{\setcounter{tocdepth}{2}}

\section{Environment Creation}





\subsection{Autonomous Reset}
We show that, given a task specification, coding agents can synthesize long-horizon reset functions by composing perception and planning APIs, including SAM3 for open-vocabulary object detection (illustrated in \Cref{fig:sam3_api}), BundleSDF for continuous 6-DOF pose tracking, and cuRobo for GPU-accelerated collision-free trajectory optimization. During execution, these reset functions consume real-time RGBD streams and proprioceptive state, which together support robust object localization, grasp pose generation, and motion planning. Coding agents can also command and read gripper torque signals, which serve as a surrogate for tactile feedback; we find that these signals are crucial for slip detection and controlling the force of the grip in precision-critical manipulation tasks.

\captionof{codesnippet}{\textbf{Tool API: SAM3 Multi-Object Segmentation.} Detects all instances of a queried object from a chosen camera view, returning per-instance masks, confidence scores, and back-projected 3D centroids.}
\label{fig:sam3_api}
\begin{appendixcode}{python}
class SegmentAllObjectsTool(Tool):
    """Detect all instances of an object using SAM3 multi-detection."""
    parameters = [
        ToolParameter("query",                "str"),
        ToolParameter("camera",               "str",   default="top"),
        ToolParameter("score_thresh",         "float", default=0.1),
        ToolParameter("image_bbox",           "list",  default=None),
        ToolParameter("image_exclude_bboxes", "list",  default=None),
        ToolParameter("filters",              "list",  default=None),
        ToolParameter("rank_by",              "dict",  default=None),
    ]
    # query:
    #     Natural-language description of the object to segment.
    # camera:
    #     Camera view to capture from. One of {top, left, right}.
    # score_thresh:
    #     Minimum SAM3 confidence score. Detections below threshold are dropped.
    # image_bbox:
    #     [x_min, y_min, x_max, y_max] pixel crop. Zeros all pixels outside the
    #     box before sending to SAM3 to suppress background noise. Returned masks
    #     remain in full-image coordinates so centroid_world_xyz back-projects correctly.
    # image_exclude_bboxes:
    #     List of [x_min, y_min, x_max, y_max] boxes to zero out before sending to
    #     SAM3. Inverse of image_bbox — used to blank known distractors (e.g. gripper
    #     jaws). Combinable with image_bbox: image_bbox is applied first, then
    #     exclusions zero out within the kept region.
    # filters:
    #     List of {metric, op, value} dicts AND-ed together to prune detections.
    #     metric in {confidence, area, AR, C, P}; op in {>=, <=}.
    #     AR: aspect ratio; C: compactness 4*pi*area/P^2; P: perimeter in pixels.
    #     Detections with None or non-finite metric values are dropped.
    # rank_by:
    #     {metric, direction} dict controlling sort order of returned detections.
    #     metric in {confidence, area, AR, C, P}; direction in {max, min}.
    #     Default: {confidence, max}.

    def execute(self, **kwargs) -> ToolResult:
        rgb         = self._fetch_camera_image(kwargs["camera"])
        rgb_for_sam = self._preprocess(rgb, kwargs["image_bbox"],
                                       kwargs["image_exclude_bboxes"])
        detections  = requests.post(f"{self._sam3_url}/segment_all",
                          json={"text":           kwargs["query"],
                                "image_b64":      to_b64(rgb_for_sam),
                                "score_threshold":kwargs["score_thresh"]}).json()
        depth, intr, T_cam_world = _fetch_camera_params(self._cam, kwargs["camera"])
        results = [
            SegmentationResult(
                mask               = np.load(io.BytesIO(base64.b64decode(d["mask_b64"]))),
                score              = d["score"],
                bbox_xywh          = d["bbox_xywh"],
                mask_area          = d["mask_area"],
                perimeter_px       = _shape_stats(mask)["perimeter_px"],
                compactness        = _shape_stats(mask)["compactness"],
                centroid_uv        = _shape_stats(mask)["centroid_uv"],
                centroid_world_xyz = _project_centroid(mask, depth, intr, T_cam_world),
                depth_m            = depth[v, u],
            ) for d in detections
        ]
        return ToolResult(success=True,
                          data=_rank_results(_apply_filters(results,
                                             kwargs["filters"]), kwargs["rank_by"]))
\end{appendixcode}

\begin{figure}[H]
    \centering
    \includegraphics[width=0.9\linewidth]{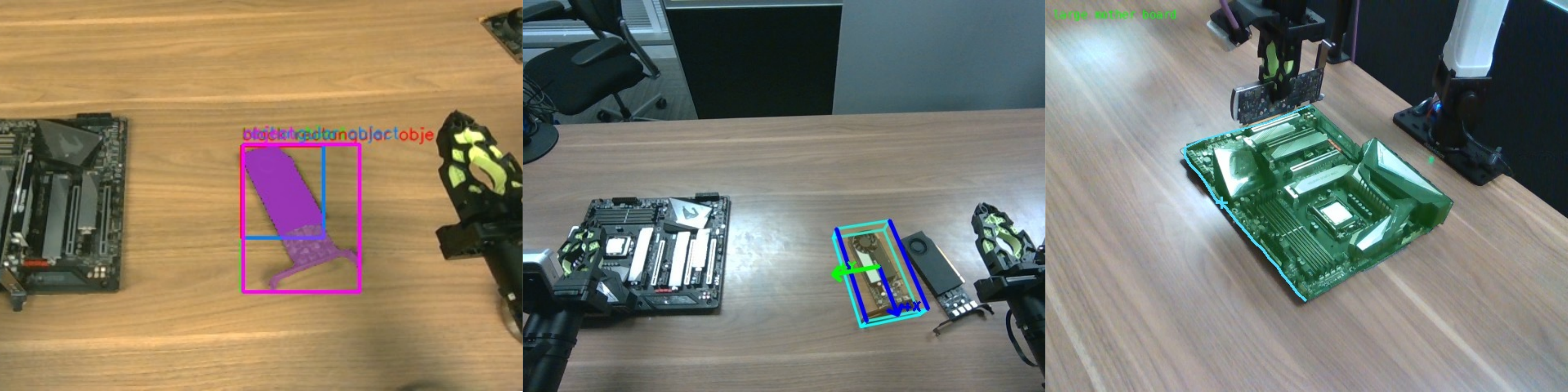}
    \caption{Visual tools for GPU insertion. Agent-written auxiliary detection tools for GPU localization via SAM3 and depth estimation.}
    \label{fig:visual_tools_gpu_insertion}
\end{figure}

\vspace{0.5cm}
From these API tools, coding agents can synthesize perception skills such as semantic segmentation and 3D bounding box generation, as well as manipulation skills including pick-and-place, object reorientation, localization, and bimanual handover. Composing these skills with failure detection and retry mechanisms yields long-horizon reset functions that are robust to real-world perturbations. \Cref{fig:gpu_reset_policy} illustrates a concrete example for the GPU insertion task, where the agent sequences SAM3-based object localization, RANSAC~\cite{ransac} and OBB~\cite{obb}-based 3D bounding box estimation, torque-verified grasping, and cuRobo collision-free handover into a complete reset pipeline. Visualizations of the intermediate segmentation and bounding box outputs are provided in \Cref{fig:visual_tools_gpu_insertion}.

\captionof{codesnippet}{\textbf{Reset Policy: GPU Insertion Recovery.} Localizes the motherboard and target slot with SAM3, grasps and extracts the GPU with a torque-verified grip, then carries it to a parking pose via a cuRobo collision-free handover.}
\label{fig:gpu_reset_policy}
\begin{appendixcode}{python}
def reset_gpu_insertion():
    go_home()

    # --- SAM3: localize motherboard and target PCIe slot ---
    scene = sam3_localize_motherboard(camera="top")
    hover_pos, scene = acquire_slot_hover(scene)

    # --- grasp GPU and extract from slot ---
    descend_guided(scene)
    close_gripper(torque_limit=LEFT_CLOSE_TORQUE_LIMIT)
    verify_grasp_or_raise()                              # gripper torque check
    lift_and_pull(scene)

    # --- handover: carry GPU to parking pose on table ---
    freespace_move(left_target_pos=TABLE_APPROACH_POS,
                   left_target_rpy=TABLE_PLACE_RPY,
                   backend="curobo")
    release_glide(TABLE_PLACE_POS, TABLE_PLACE_RPY)     # synchronized open
    cartesian_retract_after_release()
    go_home()
\end{appendixcode}

\subsection{Autoresearch-driven Reward Design} 
Autoresearch can be applied to automate the design of reward functions. This procedure requires human effort in only two steps: first, collect a few minutes of representative success and failure demonstrations; second, specify the reward design requirements, including target precision, recall, and inference latency. 

As an example, we assign the agent to build a visual reward function that detects whether a zip tie is fastened. For evaluation, we record a few minutes of labeled success and failure snapshots of zip tie fastening as a sandboxed held-out evaluation set—the agent may study failures, but cannot train on the test set or alter metric computation. The agent autonomously probes the SAM3 API via prompt search, tunes cropping and depth thresholds to suppress background noise, and meets the 150 ms latency budget by converting to ONNX, compiling the computation graph for simultaneous multi-object forward passes. The decision-making logic converges on the two-view geometric test, illustrated in \Cref{fig:zip_tie_reward}. A simplified implementation is provided in \Cref{fig:reward_impl}.

Autoresearch is not limited to visual rewards; the coding agent can compose hybrid reward functions that fuse multiple sensing modalities, including proprioception and end-effector force estimates. As an example, \Cref{fig:pin_insertion_reward} illustrates an autonomously designed reward for pin insertion that combines visual alignment, depth of insertion, and contact force signals.

\captionof{codesnippet}{\textbf{Dual-View Zip Tie Verification.} Combines the top and right camera views in a geometric test of whether the zip-tie strap passes through the head, guarding against single-view false positives.}
\label{fig:reward_impl}
\begin{appendixcode}{python}
def get_reward_from_top_right_cam(rgb_t, rgb_r):
    dets = segment_zip_tie_views(rgb_t, rgb_r, threshold=sam3_score_threshold)
    strap = union([d["mask"] for d in top2_by_score(dets["top"]["strap"])])
    head = top1_by_score(dets["top"]["head"])["mask"]
    strap = binary_dilation(strap, iterations=top_dilate)
    head = binary_dilation(head, iterations=top_dilate)
    overlap = (strap & head).sum() / head.sum()
    protrude = strap[:, :head_left_x(head)].sum() / head.sum()
    top_ok = overlap >= top_inter_frac and protrude >= top_left_mult

    right = binary_dilation(dets["right"]["strap"]["mask"], iterations=right_dilate)
    head_box = bbox_shrunk(dets["right"]["head"], frac=right_shrunk_frac)
    labeled, n_comp = label(right)
    if n_comp == 2:
        right = right | connect_components(labeled)
    right_ok = n_comp in (1, 2) and (right & head_box).any()
    return top_ok and right_ok
\end{appendixcode}

\begin{figure}[H]
    \centering
    \includegraphics[width=0.95\linewidth]{corl_2026_template_submission//figures/agent-generate-env.pdf}
    \caption{Reward function for pin insertion.
    The agent proposes a hybrid verification strategy fusing visual alignment of the
    pin tip to the hole, insertion depth from proprioception signals, and end-effector torque detection.}
    \label{fig:pin_insertion_reward}
\end{figure}

\newcommand{\api}[1]{\texttt{\detokenize{#1}}}

\section{Robot System Setup}
\label{app:robot-system}

\subsection{Fleet Architecture}
\label{app:fleet}
Our physical infrastructure is a fleet of \textbf{eight bimanual robot stations}, operated in a decentralized manner. Each station owns its robot arms and cameras, compute, and its own coding agent. Hardware-control requests from each station's agent are routed through a local FastAPI server. Coordination across stations is mediated entirely through \textbf{Git}: rather than streaming state to a central server, stations share code, configurations, tools, and results by pushing to and pulling from a common repository, so that improvements discovered on one station propagate to the rest through ordinary version control. Because all coordination flows through Git, agents can borrow ideas from one another: they freely merge changes from other branches or cherry-pick individual commits, selectively adopting promising approaches and results discovered at other stations. This keeps the fleet loosely coupled and fault-tolerant, since stations run, fail, and recover independently, while the shared Git history serves as the single source of truth for what each station has tried and learned.
\ifdefined\nvidiareleaseomitappendixfleet
\else
\begin{figure}[h]
\centering
\includegraphics[width=\linewidth, trim=0 0 0 265, clip]{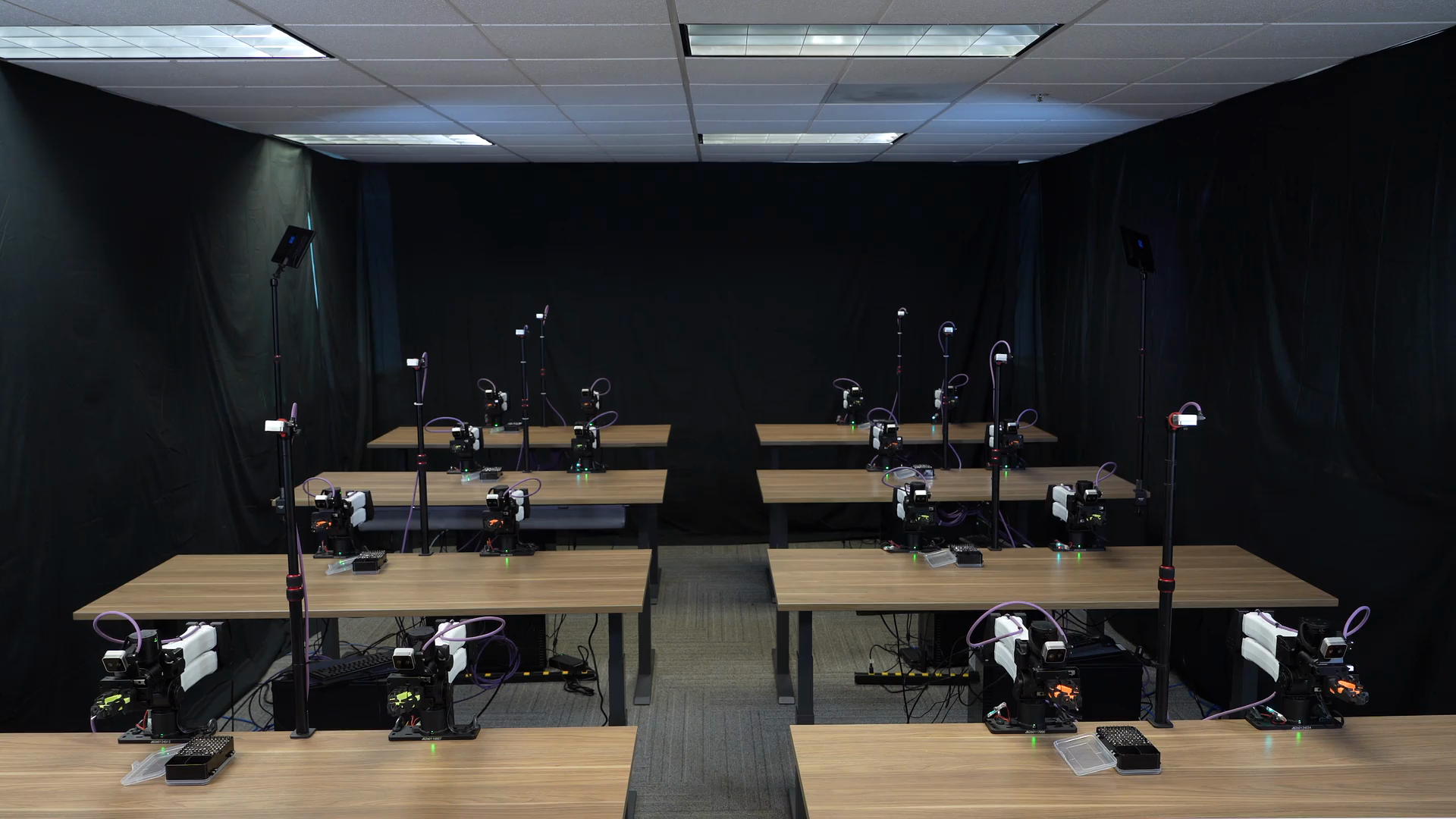}
\caption{The fleet of eight bimanual YAM robot stations. Each station owns its robot hardware, compute, and coding agent.}
\label{fig:fleet}
\end{figure}
\fi

\paragraph{Control endpoints.}
The FastAPI server exposes a small set of endpoints through which the agent drives data collection and execution: \texttt{/start} begins a rollout on the real hardware, \texttt{/restart} allocates a fresh rollout buffer directory, and \texttt{/home} returns the arms to the home configuration. The \texttt{/restart} endpoint keeps successive rollouts separate and individually addressable, so that the data from different hypotheses or experiments is never mixed together: each rollout is written to its own directory, and the agent can attribute outcomes to the specific change it was testing. These endpoints are shared across all of our tasks. The Push-T task additionally exposes \texttt{/avoid} and \texttt{/resume} endpoints to handle occlusion: \texttt{/avoid} backs the arm out of the top-camera frame so it no longer occludes the scene, and \texttt{/resume} returns the arm to its original position to continue manipulation.

\paragraph{Agent sandbox and context.}
Each station's coding agent operates within a sandbox scoped to a single autoresearch repository. Within this sandbox the agent runs with elevated autonomy: action-level permission prompts are bypassed, so it can execute commands without per-step human approval, and it is granted unrestricted internet access. The agent is provided with the complete set of robot data collected during the current autoresearch session, and is free to make full use of all available information in pursuit of its objective. For ordinary fresh sessions, raw repository state, rollout data, checkpoints, and transient logs from prior sessions are pruned before initialization, so that the agent starts from a clean, isolated workspace. For transfer experiments, such as initializing GPU insertion after pin insertion, the agent is instead given an explicit markdown summary distilled from the previous pin-insertion autoresearch session; raw trajectories, hidden logs, and checkpoints are still removed, so transfer occurs through the written summary rather than through unbounded access to prior session state. The prompt given to each station's agent is shown in \Cref{fig:autoresearch-prompt}.

\begin{figure}[t]
\begin{minted}[fontsize=\small, frame=lines, framesep=2mm, breaklines=true]{text}
/goal Achieve a 100% success rate over each window of 50 consecutive
episodes. Read @auto_research.md and fan out a subagent team to study
different algorithms (RL, BC, hybrid). You may reuse checkpoints or data
buffers across hypotheses. Create your own branch from autorl following the
instructions. Do not directly work on or push to autorl. You must actively
monitor other branches and commits from origin that branched off autorl to
leverage knowledge from those branches. You must actively push your branch to
contribute to the repo and pull from origin. Collaborate with other agents
via Git.
\end{minted}
\caption{The autoresearch prompt provided to each station's coding agent.}
\label{fig:autoresearch-prompt}
\end{figure}

\subsection{Station Hardware}
\label{app:station}
All 8 stations are hardware-identical. Each station comprises two YAM (Yet Another Manipulator) arms from I2RT in a fixed bimanual configuration, a set of cameras, and a single workstation that runs the FastAPI server, policy inference, and the station's agent.

\paragraph{Manipulators and actuation.}
Each arm is a 6-DoF manipulator equipped with a 1-DoF parallel-jaw gripper, giving seven actuated joints per arm and fourteen across the bimanual pair. All joints are driven by brushless actuators over a CAN bus. The six arm joints are run under PD control with gravity compensation, while the gripper is run in a force-limited mode that bounds grip force directly (\Cref{app:lowlevel}); we exploit this gripper-level force limiting both for robust grasping and for safe, unattended operation across the fleet.

\paragraph{Compute.}
Each station runs on a single workstation whose specifications are listed in \Cref{tab:compute}. All policy inference and on-station computation run on this single GPU, with no shared cluster or off-station compute.

\begin{table}[h]
\centering
\caption{Per-station compute specifications.}
\label{tab:compute}
\begin{tabular}{ll}
\toprule
Component & Specification \\
\midrule
GPU       & 1$\times$ NVIDIA RTX 5090, 32\,GB \\
CPU       & Intel Core Ultra 9 285K, 24 cores \\
RAM       & 128\,GB \\
OS        & Ubuntu 22.04 LTS \\
GPU stack & NVIDIA driver 595.58.03, CUDA 13.2 \\
\bottomrule
\end{tabular}
\end{table}

\paragraph{Perception.}
Unless noted otherwise, perception uses Intel RealSense \textbf{D405} cameras: one top-down camera viewing the workspace and two wrist-mounted cameras, one per arm, that provide close-range views of the grasped object and the contact region. One task also uses a side-mounted RealSense \textbf{D435i} for a wider third-person view (\Cref{app:gpu-insert}). The policy runs at \textbf{30\,Hz}, and its action targets are tracked by the low-level joint controllers described below, which run at \textbf{100\,Hz} over the CAN bus. We additionally use \textbf{Viser} for real-time, browser-based 3D visualization of robot state, camera frames, and target poses, which supports monitoring of autonomous runs, calibration debugging, and episode inspection.

\subsection{Low-Level Control}
\label{app:lowlevel}
Policy actions are inferred at \textbf{30\,Hz}, and each action target is tracked by the low-level joint controllers running at \textbf{100\,Hz} over the CAN bus. The two joint groups are controlled differently. The six arm joints use \textbf{PD control with gravity compensation}, which tracks the commanded joint targets while the feedforward gravity term offloads the static load so the PD gains only need to handle the residual error.

\vspace{0.5cm}
The 1-DoF gripper runs as a \textbf{torque-limiting compliant grasp}: instead of closing to a rigid target width, the gripper applies a bounded grip force set by a commanded torque limit. This force limiting is the key to robust and safe grasping. Because the fingers settle around the object at a bounded force rather than driving to a fixed position, the grasp accommodates variation in object pose and size and is robust to perception and placement error. The same bound caps the force the system can exert during grasping, contact, and insertion, preventing damage to the gripper, the manipulated parts, and the fixtures when an attempt is misaligned or fails. Because the fleet operates autonomously and unattended across all 8 stations, this bounded-force behavior is essential: a bad contact results in a safe stall rather than a hardware-damaging push, with no human in the loop to intervene.

\subsection{Per-Task Configuration}
\label{app:gpu-insert}

\begin{figure}[h]
\centering
\begin{minipage}[t]{0.48\linewidth}
\centering
\includegraphics[width=\linewidth]{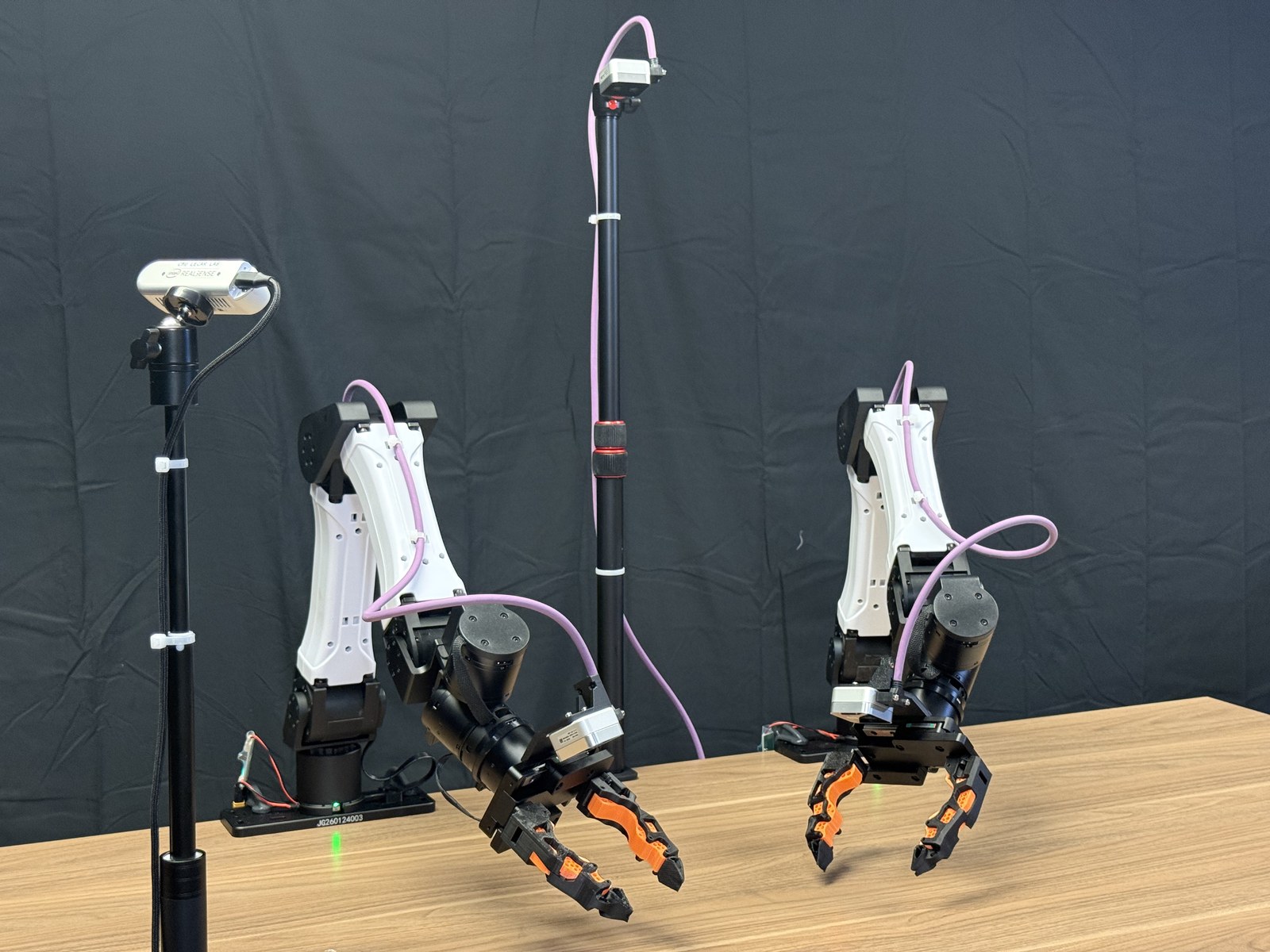}
\subcaption{Four-camera setup (GPU insertion).}
\label{fig:four-cam}
\end{minipage}\hfill
\begin{minipage}[t]{0.48\linewidth}
\centering
\includegraphics[width=\linewidth]{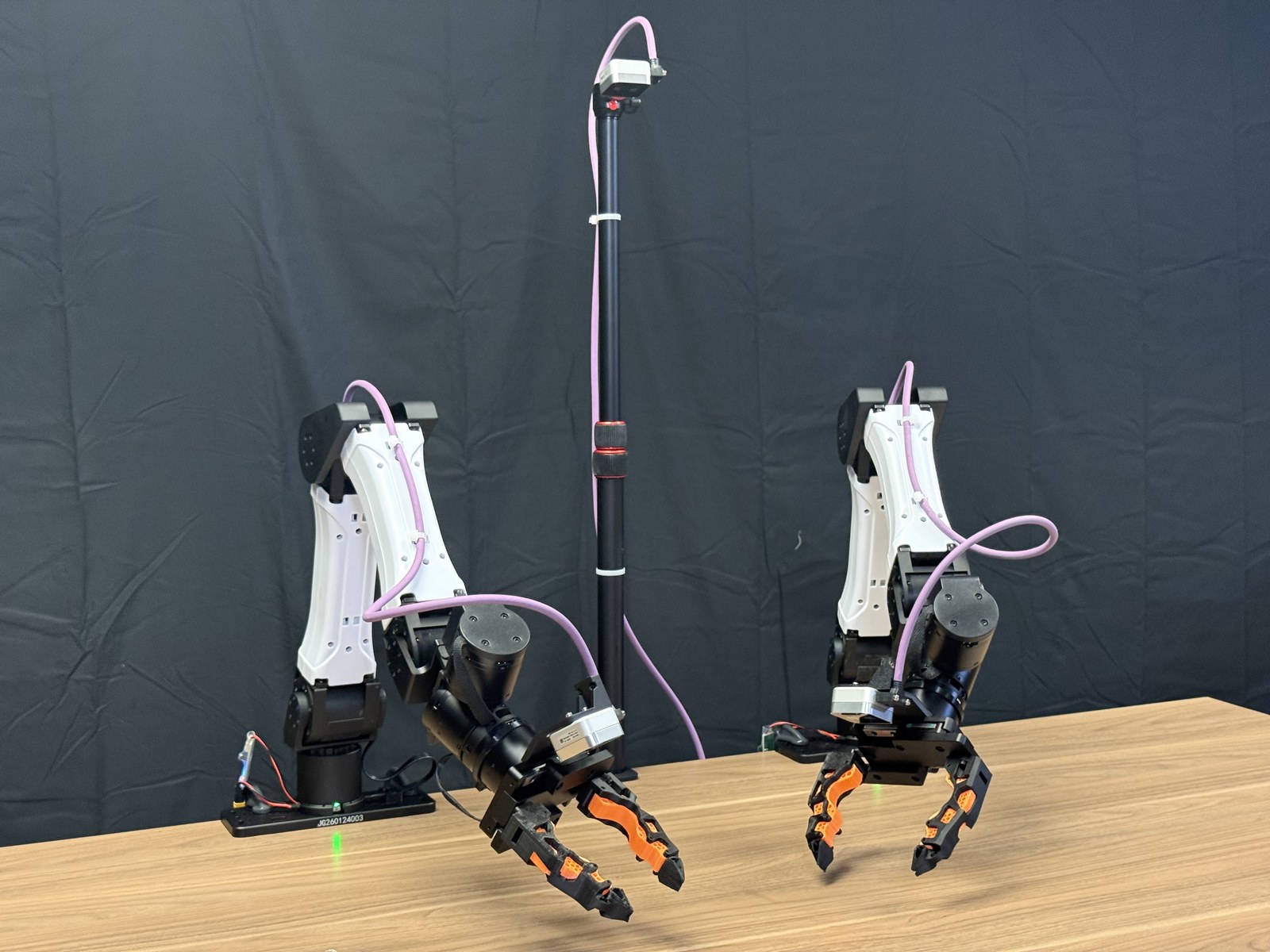}
\subcaption{Three-camera setup (pin insertion, Push-T, zip tie).}
\label{fig:three-cam}
\end{minipage}
\caption{Per-station camera configurations. All tasks use one top-down camera and two wrist-mounted cameras, all RealSense D405. The GPU insertion task in \Cref{fig:four-cam} additionally uses a side-mounted RealSense D435i on the left pole for a wider third-person view of the slot; the remaining tasks use the three-camera setup in \Cref{fig:three-cam}.}
\label{fig:camera-setups}
\end{figure}

The four tasks share the station hardware and the 30\,Hz policy loop above, and differ only in minor details of their camera and control configuration. As shown in \Cref{fig:camera-setups}, all four use one top-down camera and two wrist cameras, all RealSense D405; the GPU insertion task additionally uses a side-mounted RealSense D435i for a wider third-person view of the slot.

\subsection{Real-World RL System Integration}
We integrated the PLD-RL pipeline~\cite{pld} to provide a controlled sandbox for the coding agent to conduct algorithm auto-research with online data collection. The infrastructure is developed following the asynchronous design of SERL~\citep{luo2024serl, luo2025precise}. We implement it as a three-tier distributed system that decouples robot interaction, policy learning, and inference across separate processes. The deployment layer runs on a robot-adjacent control machine and is responsible for hardware orchestration, episode recording, and human-in-the-loop teleoperation. The second tier, learner, runs on a GPU host and trains an RL agent (actor \& critic) with pixel observations encoded through a pre-trained visual backbone. The third tier, actor, exposes a Portal/ZMQ msgpack-compatible endpoint, so that the controller can request actions through a uniform protocol shared with scripted and teleoperated control modes.

Data flow between the deployment layer and the learner is mediated by a deliberately loose disk-based contract. The deployment layer serializes each episode as a per-step observation tensor and synchronized camera streams (.mp4) into a rollout-buffer directory, accompanied by a per-step action label identifying the action source (rl, human, etc.). Meanwhile, a daemon thread (\textit{DiskBufferIngestor}) polls this directory periodically, parses newly finalized episodes, and routes transitions by action-source label following the RLPD-style data-mixing protocol~\citep{rlpd}: RL-generated transitions populate the online replay buffer, while human and manual transitions populate a separate demonstration buffer that is mixed into each training batch. This design is flexible for mixing data from different sources.

\subsection{Idea Tree for Pin Insertion}
\label{app:idea-tree-pin}
\begin{figure*}[t]
\centering
\includegraphics[width=\linewidth]{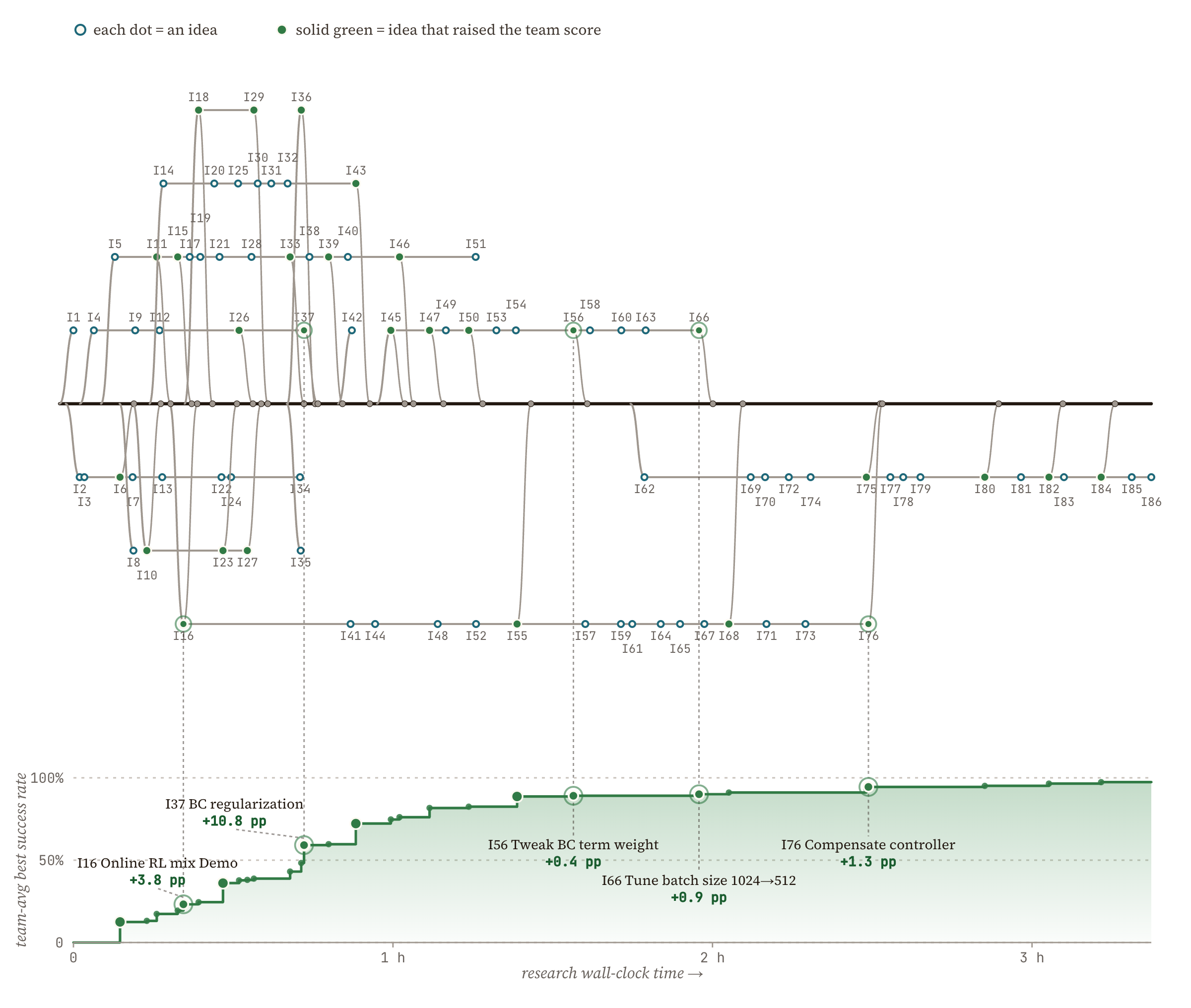}
\caption{\textbf{Idea tree and hill-climbing progress for the pin insertion task.}
\emph{Top:} the agent's idea tree, where each node $\mathrm{I}k$ is a proposed idea and
each new lane is a new idea branch. Two nodes joined by a horizontal line are related
ideas. Solid green nodes improved the team-average best success rate; hollow nodes were
evaluated but yielded no gain. The thick black line traces the lineage of the
highest-scoring idea, and circled nodes are highlighted ideas linked by dashed lines to
their milestones below. \emph{Bottom:} the team-average best success rate over research
wall-clock time, sharing the time axis of the tree. Annotated milestones report each
highlighted idea's improvement in percentage points (pp).}
\label{fig:idea-tree-pin}
\end{figure*}

\Cref{fig:idea-tree-pin} visualizes an agent-team autoresearch run on the pin
insertion task as an idea tree (top) paired with its hill-climbing curve (bottom).
Each node $\mathrm{I}k$ is an idea explored by the team; ideas that are related are
connected by a horizontal line, and each new lane corresponds to a new branch of ideas.
Solid green nodes mark ideas that improved the team-average best success rate, while
hollow nodes were evaluated but yielded no gain. The thick black line traces the lineage
of the highest-scoring idea. Circled nodes are highlighted ideas that are also annotated
on the hill-climbing curve below, to which they are connected by dashed lines. The bottom
panel reports the team-average best success rate as a function of research wall-clock time, sharing the time axis of the tree above. As the agents collaborate through Git,
a few high-impact ideas account for most of the progress---most notably BC regularization
(I37, $+10.8$\,pp)---while later ideas such as batch-size tuning (I66, $+0.9$\,pp) and
controller compensation (I76, $+1.3$\,pp) contribute smaller, incremental refinements as
the success rate approaches $100\%$.

\section{Physical Autoresearch: Ablation Studies}

\subsection{Experimental Setup}
We constructed a simplified real-world Push-T environment, \texttt{pusht-simple}, for controlled ablations of the autoresearch agent. The task preserves the core visual servoing and contact-planning structure of Push-T while reducing episode length and setup variability, making it suitable for repeated comparisons across model, harness, and visual-grounding conditions. Compared with the Push-T task in the main paper, this variant uses a relaxed success boundary so that ablations can focus on controller discovery rather than rare terminal precision. An episode is counted as successful when the pushed T block reaches the target region within the prescribed translational tolerance and its orientation is within $10^\circ$ of the goal orientation. Operationally, the agent must write a controller that rotates the T block by approximately $180^\circ$ and places it inside the enlarged success region.

\subsection{Token Utilization}

\begin{figure}
    \centering
    \includegraphics[width=1\linewidth]{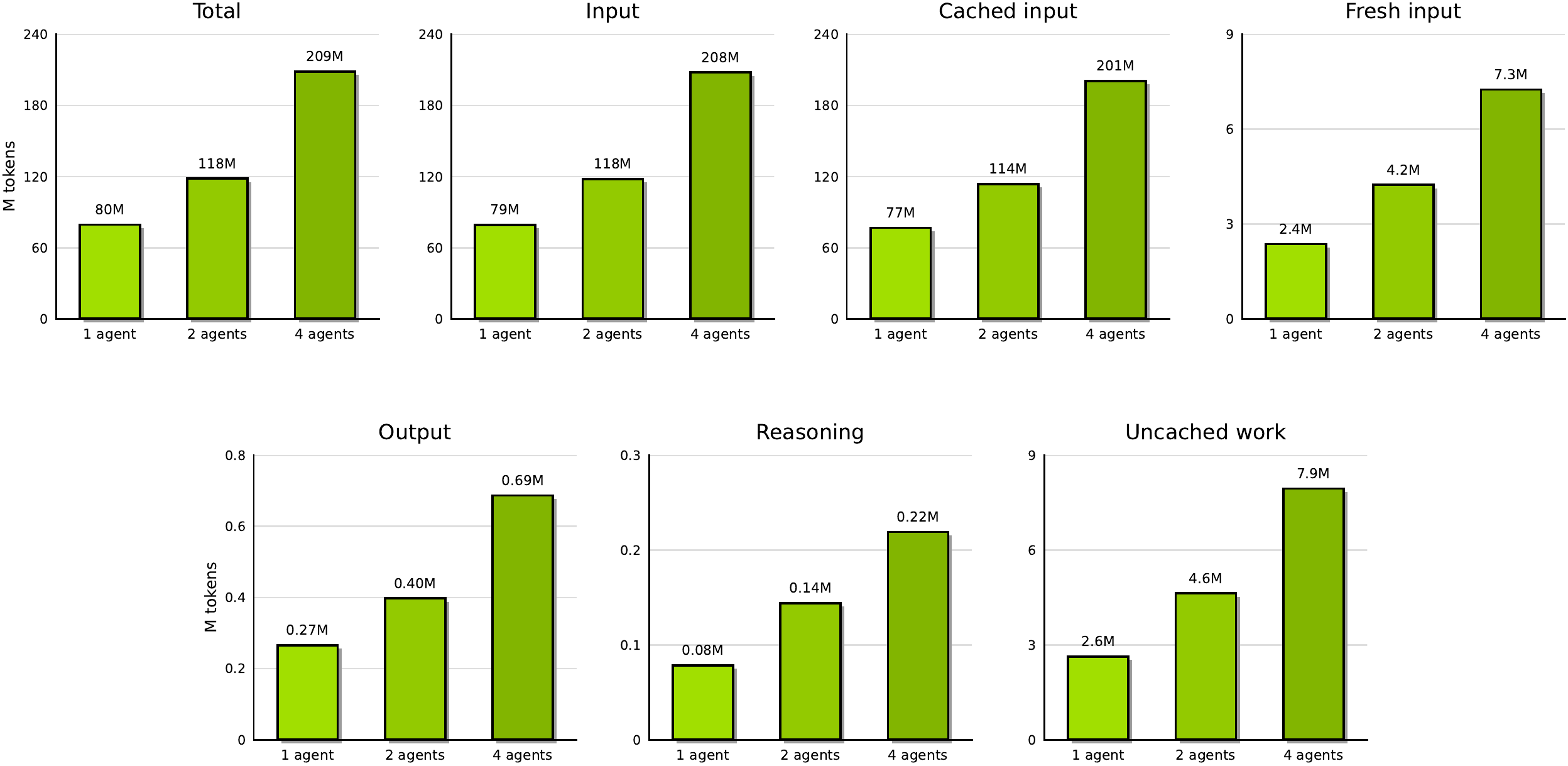}
    \caption{Token-utilization breakdown for Codex agents on the simplified real-world Push-T task. Each panel reports a different view of token consumption over the autoresearch process, separating prompt/input, generated/output, and cached-context usage where available. The figure complements the aggregate MTU analysis in the main paper by showing how token demand is distributed across planning, code editing, log inspection, and repeated policy-evaluation iterations.}
    \label{fig:token_utilization_breakdown}
\end{figure}

In the main paper, we summarize token scaling by aggregating input and output tokens. Because input, output, and cached tokens differ in cost, latency, and the type of agent activity they reflect, we additionally report token usage by category. This breakdown clarifies how much of the autoresearch budget is spent on context ingestion, code and plan generation, and reuse of cached interaction history. For a more detailed qualitative analysis, we visualize total token usage, input token usage, cached input token usage, fresh input token usage, output token usage, and reasoning token usage.

\subsection{Native Vision Capability}

\begin{figure}
    \centering
    \includegraphics[width=1.0\linewidth]{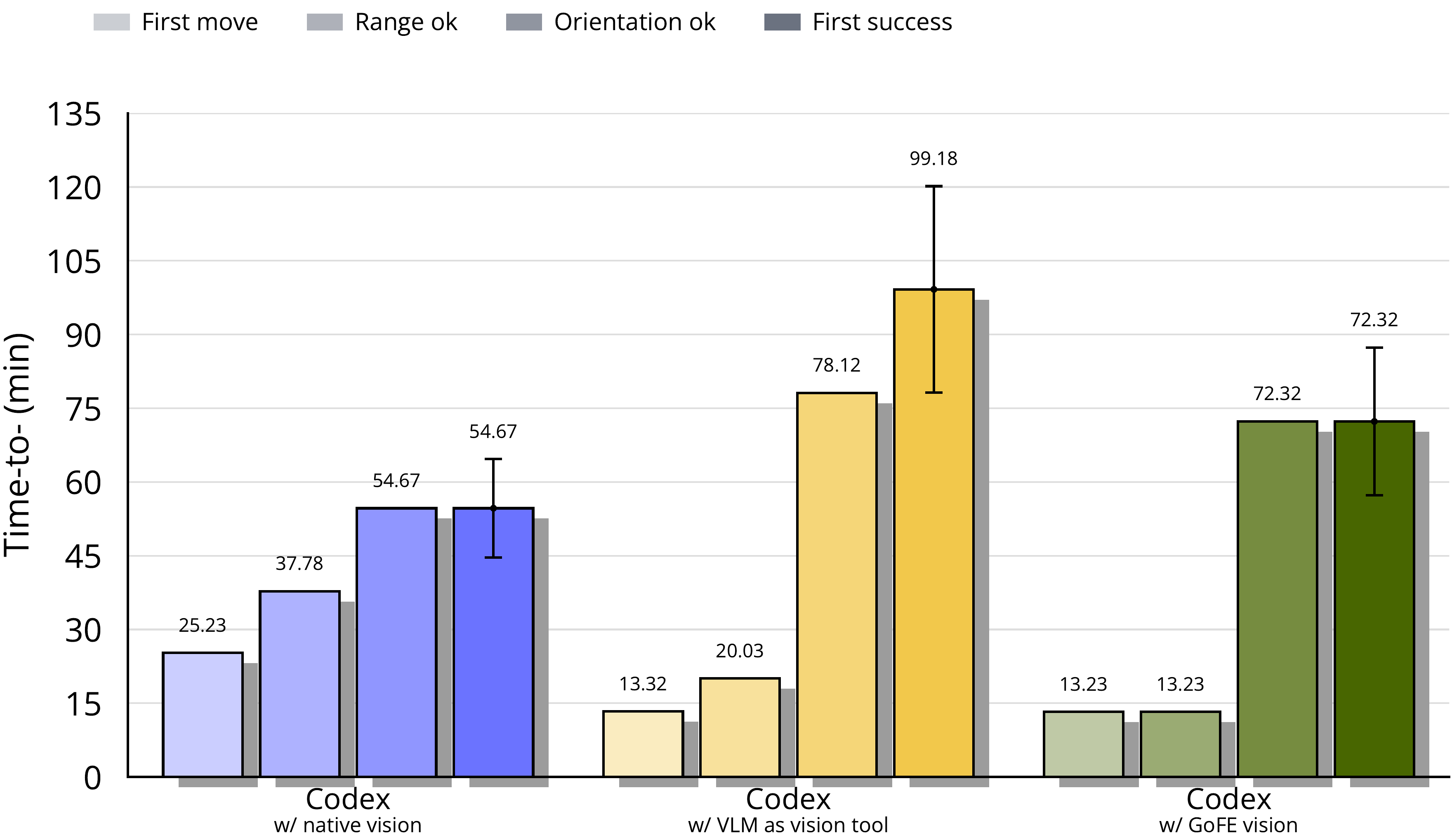}
    \caption{Stage-wise progress on the simplified real-world Push-T task under different visual-grounding conditions.}
    \label{fig:pusht_stage_progress}
\end{figure}

Since Push-T requires the agent to localize the object, infer contact geometry, and diagnose failures from evaluation rollouts, we ablate the effect of native vision capability in the coding agent. The Codex harness supports native visual understanding, meaning that the coding agent can directly inspect images. However, many open-source agent harnesses expose only text input. This experiment measures how visual access affects auto-search hill-climbing on the simplified Push-T task. In addition to the original Codex configuration, we evaluated two baselines:

\begin{enumerate}[leftmargin=1.5em]
\item \textbf{Codex without native vision;} We mask image tokens from the coding agent but provide a separate image-understanding module as a callable function. The module reads images, produces descriptions, and answers visual questions; the system prompt is modified to allow the coding agent to call this function when visual information is needed.
\item \textbf{Codex without visual capability.} We remove both native image streaming and visual function calling. In this setting, Codex can only analyze text-based information or write code to extract information from images.
\end{enumerate}

Codex with native vision reaches success first. Surprisingly, the no-vision baseline succeeds before the function-call vision baseline. This suggests that even without direct visual access, the coding agent can infer useful task state from other logging signals, while repeated image function calls introduce additional overhead during task solving.

\begin{figure}
    \centering
    \includegraphics[width=0.5\linewidth]{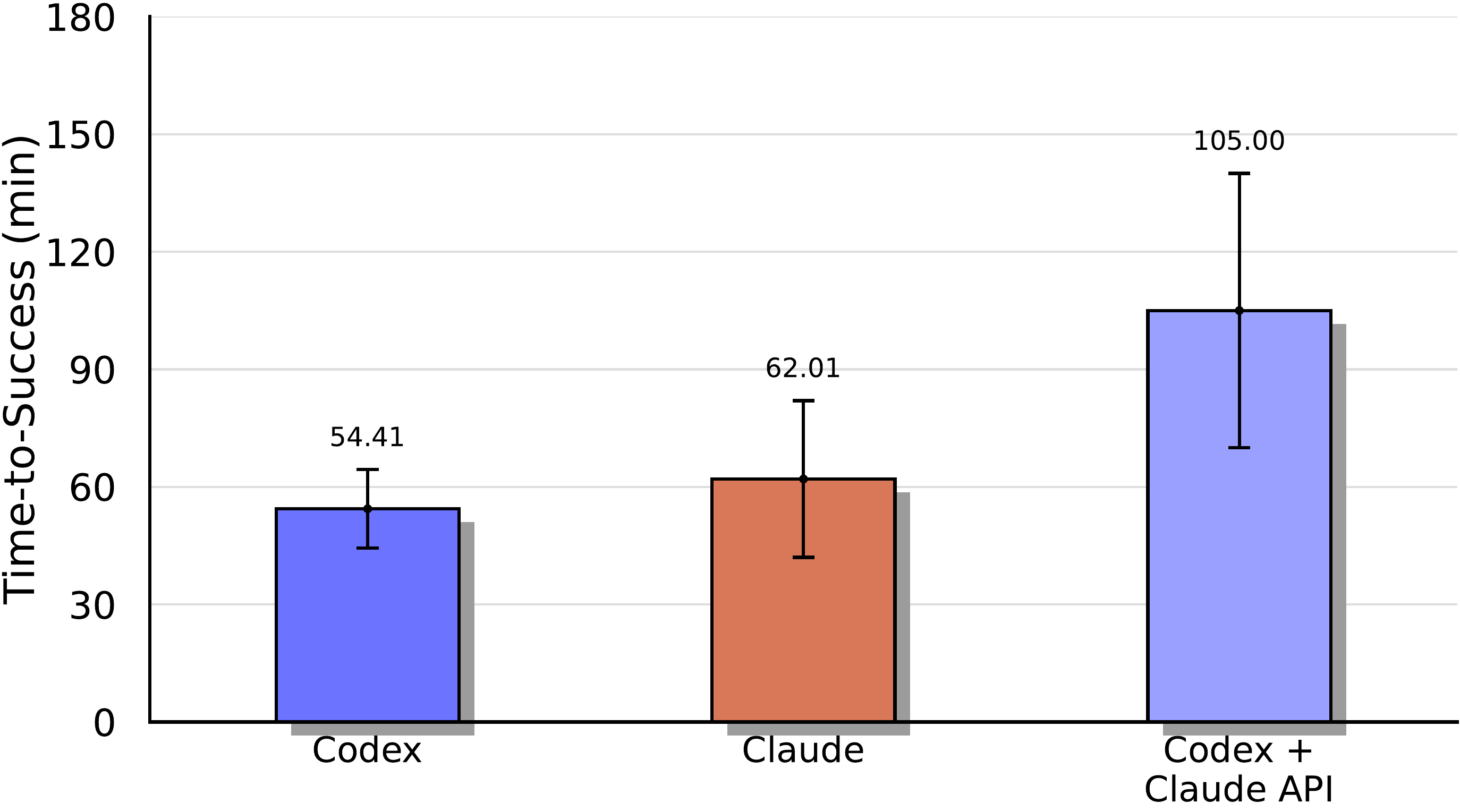}
    \caption{Time-to-success comparison across model and agent-harness configurations on the simplified real-world Push-T task. Each configuration is evaluated by the wall-clock time required to reach the predefined success criterion, capturing both model reasoning capability and the practical overhead introduced by the coding interface, tool access, execution loop, and log-inspection workflow.}
    \label{fig:pusht_time_to_success}
\end{figure}

\subsection{Model and Harness Comparison}

We study how the choice of model and agent harness affects end-to-end autoresearch performance. We compare Codex with GPT-5.5, Claude with Opus 4.7, and Codex with Opus 4.7. Codex solves the task fastest, while the Codex harness using the Opus API is the least efficient among the tested configurations.

\section{Simulation Benchmark Result}
\subsection{RoboCasa Simulation Interface and API Surface}
\label{subsec:robocasa-api-summary}

The RoboCasa365 benchmark provides the simulation-side evaluation setting for generated vision-language control scripts. We build middle-level vision and planning stacks on top of the RoboCasa simulation, and encapsulate modular and end-to-end tools as APIs. This interface is injected into generated vision scripts as a single Python namespace. For canonical evaluation scripts, APIs that leak privileged state or enable repeated retries are removed from the runtime namespace. In particular, the environment disables APIs such as \api{get_task_info}, \api{reset_env}, \api{reset_to_initial}, task-specific CLI helpers, and
oracle-target queries. The oracle target API is only exposed outside the canonical script
runtime. \Cref{tab:robocasa-api} summarizes the surface and marks, for each call, whether it is
available in the canonical script runtime (\checkmark) or gated to CLI/oracle modes
(\textsf{C}/\textsf{O}).

{\footnotesize
\setlength{\tabcolsep}{4pt}
\renewcommand{\arraystretch}{1.15}
\begin{longtable}{@{}l p{7.2cm} c@{}}
\caption{RoboCasa environment API surface used in our evaluation. \emph{Avail.}\
indicates availability in the canonical script runtime: \checkmark{} available;
\textsf{C} exposed only in non-canonical CLI mode; \textsf{O} exposed only when the Oracle interface is explicitly enabled.}
\label{tab:robocasa-api}\\
\toprule
\textbf{API} & \textbf{Description} & \textbf{Avail.}\\
\midrule
\endfirsthead
\toprule
\textbf{API} & \textbf{Description} & \textbf{Avail.}\\
\midrule
\endhead
\bottomrule
\endlastfoot

\multicolumn{3}{@{}l}{\textit{State and gripper}}\\
\addlinespace[2pt]
\api{get_robot_state}   & Joint positions, EEF pose, gripper measurement, and base/arm state & \checkmark\\
\api{get_gripper_info}  & Gripper command, measured state, contact bodies, and actuator force & \checkmark\\
\api{set_gripper}       & Set normalized gripper command ($1$ open, $0$ closed) & \checkmark\\
\api{open_gripper}      & Open the gripper & \checkmark\\
\api{close_gripper}     & Close the gripper & \checkmark\\
\addlinespace[3pt]
\multicolumn{3}{@{}l}{\textit{Motion}}\\
\addlinespace[2pt]
\api{move_with_curobo}            & cuRobo plan, OSC waypoint tracking, scene collision & \checkmark\\
\api{move_with_pyroki}            & Pyroki IK with SE(3) interpolation, no scene collision & \checkmark\\
\api{move_with_curobo_joint_exec} & cuRobo plan executed in joint-position mode & \checkmark\\
\api{move_with_pyroki_joint_exec} & Pyroki plan executed in joint-position mode & \checkmark\\
\api{freespace_move}              & Backward-compatible wrapper over configured backend & \checkmark\\
\api{move_eef}                    & Move to a direct end-effector target & \checkmark\\
\api{select_best_grasp}           & Planner-check grasp candidates and select a feasible one & \checkmark\\
\api{go_home}                     & Return to the home configuration & \checkmark\\
\api{nudge}                       & Small local end-effector displacement & \checkmark\\
\api{nudge_brutal}                & Small local move with collision avoidance off & \checkmark\\
\api{rotate_gripper_in_place}     & In-place wrist rotation & \checkmark\\
\addlinespace[3pt]

\multicolumn{3}{@{}l}{\textit{Camera and visualization}}\\
\addlinespace[2pt]
\api{get_camera_image_480_640}      & RGB image at $480\times640$ & \checkmark\\
\api{get_camera_depth_480_640}      & Aligned depth map at $480\times640$ & \checkmark\\
\api{get_camera_intrinsics_480_640} & Intrinsics matched to $480\times640$ & \checkmark\\
\api{get_camera_image}              & Legacy RGB image alias at the environment-native resolution & \checkmark\\
\api{get_camera_depth}              & Legacy depth alias at the environment-native resolution & \checkmark\\
\api{get_camera_intrinsics}         & Legacy intrinsics alias at the environment-native resolution & \checkmark\\
\api{get_camera_extrinsics}         & Camera extrinsics & \checkmark\\
\api{set_debug_markers}             & Place visualization markers & \checkmark\\
\addlinespace[3pt]

\multicolumn{3}{@{}l}{\textit{Perception}}\\
\addlinespace[2pt]
\api{detect_object}               & Text-conditioned object detection & \checkmark\\
\api{segment_object}              & Single-instance segmentation & \checkmark\\
\api{segment_object_all}          & Multi-instance segmentation & \checkmark\\
\api{detect_objects_oneshot}      & One-shot object detection & \checkmark\\
\api{sample_grasp_pose_anygrasp}  & AnyGrasp grasp-pose sampling & \checkmark\\
\api{vlm_query}                   & Vision-language model query & \checkmark\\
\addlinespace[3pt]

\multicolumn{3}{@{}l}{\textit{Planner and collision}}\\
\addlinespace[2pt]
\api{update_planner_world}    & Set the cuRobo collision geometry & \checkmark\\
\api{refresh_planner_world}   & Refresh collision geometry from the scene & \checkmark\\
\api{disable_collision_avoid} & Disable collision avoidance & \checkmark\\
\addlinespace[3pt]

\multicolumn{3}{@{}l}{\textit{Task, policy, and navigation}}\\
\addlinespace[2pt]
\api{get_task_description}    & Natural-language task instruction & \checkmark\\
\api{use_policy_output}       & Query an external policy's output & \checkmark\\
\api{get_base_state}          & Mobile-base state & \checkmark\\
\api{execute_base_trajectory} & Execute a planned base trajectory & \checkmark\\
\api{create_goat_agent}       & Instantiate a GOAT navigation agent & \checkmark\\
\api{set_goat_goal}           & Set the GOAT navigation goal & \checkmark\\
\api{goat_navigate}           & Run GOAT navigation & \checkmark\\
\api{get_goat_state}          & Return GOAT state and configuration & \checkmark\\
\api{reset_goat}              & Reset GOAT state & \checkmark\\
\addlinespace[3pt]

\multicolumn{3}{@{}l}{\textit{Utilities}}\\
\addlinespace[2pt]
\api{display_rpy_to_quat} & Convert display RPY angles to an \api{xyzw} quaternion & \checkmark\\
\api{np}, \api{numpy}      & NumPy convenience exports for generated scripts & \checkmark\\
\addlinespace[3pt]

\multicolumn{3}{@{}l}{\textit{Gated / removed in the canonical script runtime}}\\
\addlinespace[2pt]
\api{get_task_info}      & Structured task information & \textsf{C}\\
\api{reset_env}          & Reset the environment & \textsf{C}\\
\api{reset_to_initial}   & Reset to the initial state & \textsf{C}\\
\api{get_oracle_targets} & Ground-truth target pose / oracle query & \textsf{O}\\
\end{longtable}
}

\paragraph{Camera and visualization.}
The camera APIs provide RGB-D observations, camera intrinsics, and extrinsics. We use resolution-specific calls to keep RGB, depth, and intrinsics mutually aligned, while retaining unsuffixed aliases for backward compatibility. Debug markers are used only for visualizing intermediate targets during development.

\paragraph{Motion.}
The motion APIs provide both cuRobo- and Pyroki-based execution. cuRobo uses scene collision geometry and is the default choice for free-space manipulation, whereas Pyroki provides lightweight SE(3) interpolation with inverse kinematics but no scene-level collision model. The standard mode
tracks end-effector waypoints with OSC, while joint-execution variants are used when faithfully following the planned joint path is important, such as near joint limits or tight obstacles.

\paragraph{Perception.}
The perception APIs provide text-conditioned detection, segmentation, one-shot detection,
grasp-pose sampling, and vision-language queries. These calls allow generated scripts to ground
language instructions in visual observations without accessing the simulator ground truth.

\paragraph{Planner and collision.}
The planner APIs maintain the cuRobo collision geometry from the simulated scene. Collision
avoidance is enabled by default for ordinary free-space motion, and collision disabling is reserved
for short, intentional contact-rich actions or known collision-model artifacts.

\subsection{RoboCasa Evaluation Protocol}
\label{subsec:robocasa-eval-protocol}

We evaluate each RoboCasa task in simulation using the same environment wrapper
and script runtime described above. Each script generated by the coding agent is executed
once per episode through the canonical evaluation harness. The harness creates
the RoboCasa environment, injects the restricted Python tool namespace into the
script, runs the script to completion, records videos and logs, and then queries
the environment's native success function for the final score. We report binary
task success as defined by RoboCasa's native success predication: an episode is
counted as successful when this predicate returns true at the end of execution.

\paragraph{Fixed seeds and layouts.}
For each reported 40-episode evaluation, we use a fixed list of seed triplets.
Each triplet contains a RoboCasa random seed, a layout identifier, and a style
identifier: \texttt{\detokenize{(seed, layout_id, style_id)}}.
The 40 triplets are generated once using a deterministic seed-list generator
with generator seed 42, and then saved as a static evaluation file. For all
reported evaluations, including smaller diagnostic subsets, we use rows from
this saved file rather than regenerating a new list, so that subsets are
prefixes or explicit selections from the same canonical 40-episode benchmark.
The same triplet list is used for all methods compared on a task.

\paragraph{Fair comparison with GR00T.}
When comparing generated vision scripts with GR00T, we use the same task name,
same seed triplets, same initial simulator states, same camera configuration,
and same RoboCasa success predicate. GR00T and the vision scripts are therefore
scored on matched episodes rather than on separately sampled scenes. The task
language prompt is the same RoboCasa instruction, and neither method receives
oracle target poses or simulator object states during execution. The methods
differ in their execution pathway: GR00T acts through its learned policy
rollout, while the generated scripts compose the public perception, planning,
and manipulation APIs listed in \Cref{tab:robocasa-api}. We report success over the same fixed evaluation set so that differences reflect the policy or script behavior rather than different scene sampling or privileged information.

\paragraph{Discovered policy demonstration.}
In addition to the matched-seed success rates reported in the main paper, the
project website at \website includes qualitative
demonstrations of the coding-agent-discovered policy. In particular, the videos
show the agent-discovered strategy of combining detection and motion planning to
move the gripper to a hover pose above the target object before grasping, which
is the RoboCasa strategy transferred to the real-world scissors and zip tie
task in the main paper.

\begin{figure}
    \centering
    \includegraphics[width=0.8\linewidth]{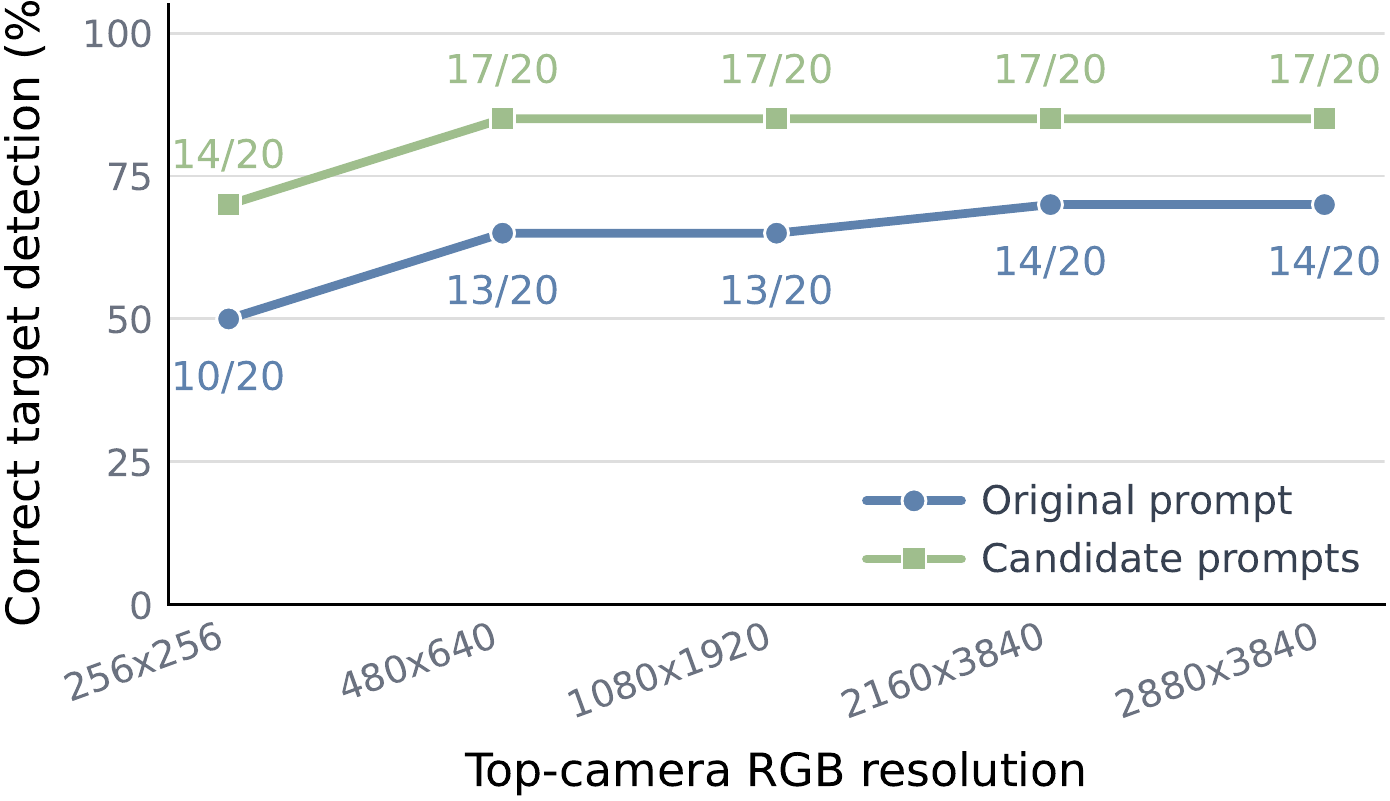}
    \caption{SAM3 target-object detection accuracy on RoboCasa counter-to-cabinet scenes as a function of top-camera RGB resolution. Each point evaluates 20 target-object queries. The original task prompt improves from 10/20 correct detections at $256\times256$ to 14/20 at higher resolutions, while candidate prompts generated by the coding agent improve from 14/20 to 17/20 and then plateau.}
    \label{fig:robocasa_sam3_resolution}
\end{figure}

\begin{figure}
    \centering
    \includegraphics[width=0.9\linewidth]{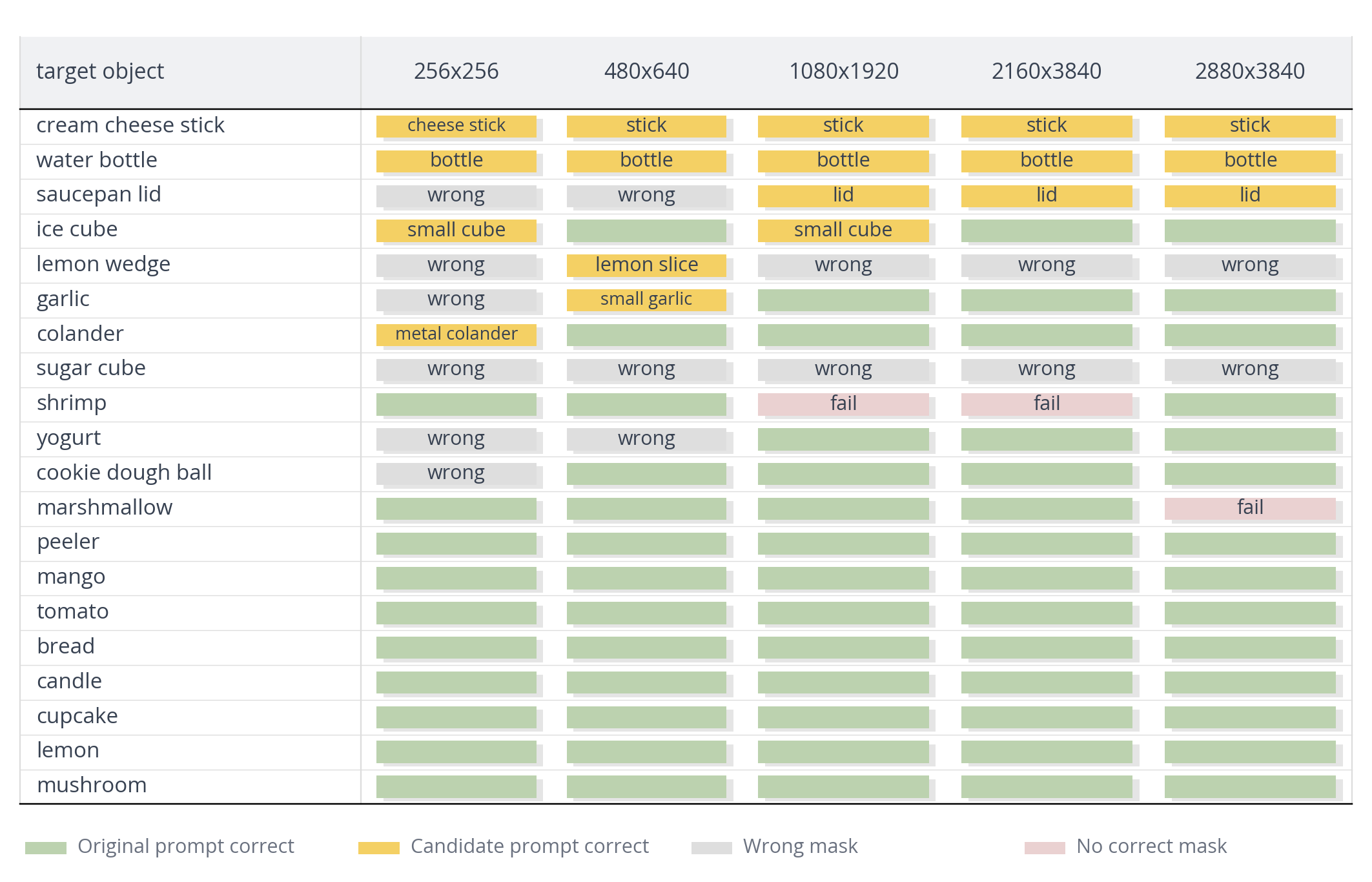}
    \caption{Per-object SAM3 detection outcomes across top-camera resolutions and prompt variants for the RoboCasa counter-to-cabinet diagnostic set. Green cells indicate that the original target prompt produced a correct mask, yellow cells indicate that a candidate prompt produced a correct mask, gray cells indicate an incorrect mask, and red cells indicate that no correct mask was returned.}
    \label{fig:robocasa_sam3_prompt_table}
\end{figure}

\begin{figure}
    \centering
    \includegraphics[width=1\linewidth]{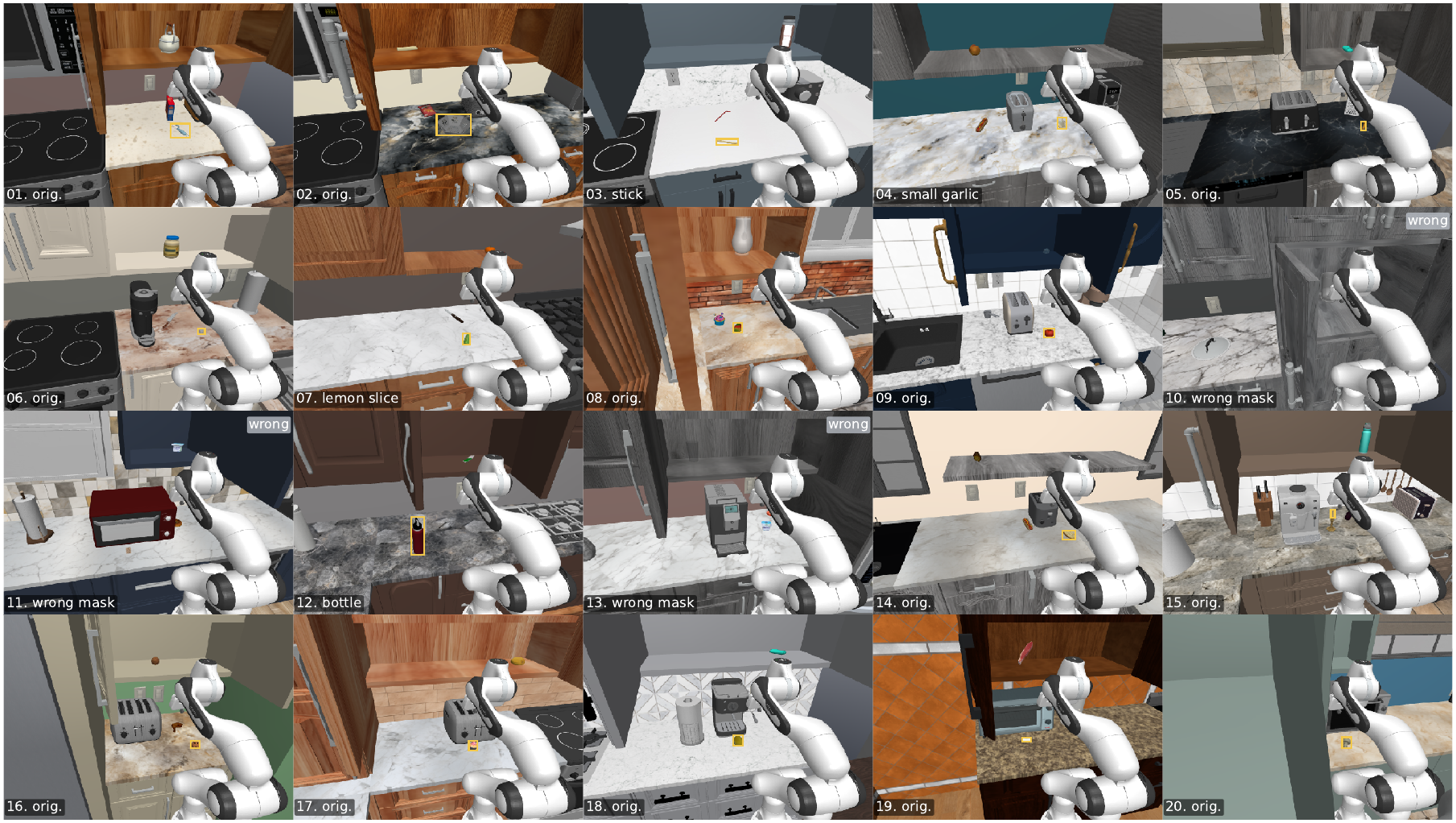}
    \caption{Qualitative SAM3 mask outputs for 20 RoboCasa counter-to-cabinet target objects at $480\times640$ resolution. Yellow boxes show selected target masks; labels indicate whether the original prompt, a candidate prompt, or a wrong mask was used.}
    \label{fig:robocasa_sam3_bbox_grid}
\end{figure}

\paragraph{Result Analysis}
We identified a perception bottleneck in which SAM3 can return an incorrect mask or no usable mask for small or ambiguous RoboCasa objects. \Cref{fig:robocasa_sam3_resolution,fig:robocasa_sam3_bbox_grid} isolate this effect by varying image resolution and prompt wording on the same counter-to-cabinet diagnostic set. Increasing the top-camera resolution from $256\times256$ to $480\times640$ improves detection, but resolution alone does not solve all failures. Allowing the coding agent to rewrite object prompts further improves target detection and recovers several cases where the original task wording produces a distractor mask. This analysis shows that generated RoboCasa scripts are limited not only by planning and control, but also by the reliability of the perception API exposed to the agent; prompt search and resolution selection are therefore useful parts of the agent's simulation-side tool use.

\end{document}